\documentclass[11pt]{article}

\usepackage[margin=1.2in]{geometry}

\usepackage{microtype}
\usepackage{graphicx}
\usepackage{subcaption}
\usepackage{booktabs} % for professional tables

\usepackage{natbib}

\usepackage{hyperref}

\usepackage{authblk}

\usepackage{amsmath}
\usepackage{amssymb}
\usepackage{mathtools}
\usepackage{amsthm}

\usepackage[capitalize,noabbrev]{cleveref}

\usepackage[textsize=tiny]{todonotes}

\usepackage{xspace}
\newtheorem{theorem}{\bf{Theorem}}

\newtheorem{proposition}{\bf{Proposition}}

\newtheorem{corollary}{Corollary}
\newtheorem{definition}{\bf{Definition}}
\newtheorem{assumption}{\bf{Assumption}}
\newtheorem{remark}{\bf{Remark}}
\newtheorem{example}{\bf{Example}}
\def\QED{~\rule[-1pt]{5pt}{5pt}\par\medskip}

\newenvironment{theorem*}[1][]{%
    \par\addvspace{\topsep}% Space above
    \noindent{\bfseries Theorem\ifx&#1&\relax\else\ (#1)\fi\quad}\itshape% Theorem header with optional title
    \ignorespaces% Remove space after theorem header
}{%
    \par\addvspace{\topsep}% Space below
}

\newenvironment{lemma*}[1][]{%
    \par\addvspace{\topsep}% Space above
    \noindent{\bfseries Lemma\ifx&#1&\relax\else\ (#1)\fi\quad}\itshape% Lemma header with optional title
    \ignorespaces% Remove space after lemma header
}{%
    \par\addvspace{\topsep}% Space below
}

\newenvironment{proposition*}[1][]{%
    \par\addvspace{\topsep}% Space above
    \noindent{\bfseries Proposition\ifx&#1&\relax\else\ (#1)\fi\quad}\itshape% Lemma header with optional title
    \ignorespaces% Remove space after lemma header
}{%
    \par\addvspace{\topsep}% Space below
}

\newenvironment{corollary*}[1][]{%
    \par\addvspace{\topsep}% Space above
    \noindent{\bfseries Corollary\ifx&#1&\relax\else\ (#1)\fi\quad}\itshape% Lemma header with optional title
    \ignorespaces% Remove space after lemma header
    }{%
    \par\addvspace{\topsep}% Space below
}

\newenvironment{assumption*}[1][]{%
    \par\addvspace{\topsep}% Space above
    \noindent{\bfseries Assumption\ifx&#1&\relax\else\ (#1)\fi\quad}\itshape% Assumption header with optional title
    \ignorespaces% Remove space after assumption header
}{%
    \par\addvspace{\topsep}% Space below
}

\newenvironment{remark*}[1][]{%
    \par\addvspace{\topsep}% Space above
    \noindent{\bfseries Remark\ifx&#1&\relax\else\ (#1)\fi\quad}\itshape% Lemma header with optional title
    \ignorespaces% Remove space after lemma header
}{%
    \par\addvspace{\topsep}% Space below
}

\DeclareMathAlphabet{\altmathcal}{OMS}{cmsy}{m}{n} % cursive letters that are not too fancy

\newcommand{\argmin}[1]{\underset{#1}{\operatorname{arg \ min \ }}}
\newcommand{\argmax}[1]{\underset{#1}{\operatorname{arg \ max \ }}}

\newcommand{\mc}{\mathcal}

\newcommand{\bb}{\mathbb}

\newcommand{\indicator}[1]{\mathbbm 1_{#1}}

\newcommand{\prob}[1]{\bb P\left( #1 \right)}
\newcommand{\expect}[2]{\bb E_{#1}\left[ #2 \right]}

\renewcommand{\arraystretch}{1}

\renewcommand{\mathbf}{\boldsymbol}

\newcommand{\fx}{f^*_{\mathrm{pool}}}
\newcommand{\fxd}{f^*_{\mathrm{full}}}
\newcommand{\fxm}{f^*_{\mathrm{DG}}}
\newcommand{\fxi}{f^*_{\mathrm{DI}}}

\newcommand{\fxg}{f^*_{\mathrm{pool},\, \mc{G}}}

\newcommand{\fxhat}{\widehat{f}_{\mathrm{pool}}}

\newcommand{\fxihat}{\widehat{f}_{\mathrm{DI}}}

\newcommand{\rx}{R^*_{\mathrm{pool}}}
\newcommand{\rxd}{R^*_{\mathrm{full}}}

\newcommand{\rxi}{R^*_{\mathrm{DI}}}

\newcommand{\rxg}{R^*_{\mathrm{pool},\, \mc{G}}}

\newcommand{\rxif}{R^*_{\mathrm{DI},\, \mc{F}}}

\usepackage{enumitem}
\usepackage{bbm}
\usepackage{multirow}
\usepackage{tablefootnote}
\usepackage{bm}
\usepackage{makecell} 
\usepackage{ragged2e}  % for \raggedright in tabular environments
\usepackage{array}     % for \arraybackslash
\usepackage{tcolorbox}

\usepackage{colortbl}
\newcommand{\graymidrule}{\arrayrulecolor{gray!80}\midrule\arrayrulecolor{black}}

\title{\textbf{Domain Generalization Under Posterior Drift}}

\author[1]{Yilun Zhu\thanks{Corresponding authors: allanzhu@umich.edu, clayscot@umich.edu}}
\author[1]{Naihao Deng}
\author[2]{Naichen Shi}
\author[3]{Aditya Gangrade}
\author[1]{Clayton Scott}

\affil[1]{University of Michigan}
\affil[2]{Northwestern University}
\affil[3]{Boston University}

\date{}

\begin{document}
\maketitle

% \vspace{-2em}

\begin{abstract}
    Domain generalization (DG) is the problem of generalizing from several distributions (or domains), for which labeled training data are available, to a new test domain for which no labeled data is available. 
    For the prevailing benchmark datasets in DG, 
    there exists a single classifier that performs well across all domains. 

    In this work, we study a fundamentally different regime where the domains satisfy a \emph{posterior drift} assumption, in which the optimal classifier might vary substantially with domain. We establish a decision-theoretic framework for DG under posterior drift, and investigate the practical implications of this framework through experiments on language and vision tasks. 
    % ``domain-informed ERM,'' wherein feature vectors are augmented with domain-specific information, outperforms pooling-ERM. These claims are supported by a theoretical framework and experiments on language and vision tasks.

    % We argue that domain-agnostic strategies like IFL are theoretically suboptimal in these regimes. We establish a framework for domain-adaptive classification that explicitly leverages domain metadata, demonstrating that domain-informed predictors significantly outperform invariant approaches on tasks with posterior drift.
\end{abstract}

\section{Introduction}

Domain generalization (DG) is the learning problem where the learner has access to labeled data from several source domains, and the goal is to generalize to a new target domain for which no labeled data is available. %Letting 
%\( X \) denote the input features, \( Y \) the label, and 
%\( D \) denote the domain index, 
More precisely, DG assumes labeled training data from each of several training domains $D_1, \ldots, D_N$, and the goal is to produce a classifier that works on any test domain $D$ from which no labeled data are available.

DG was first formulated by \citet{blanchard2011generalizing}. In this work and the follow-up work of \citet{muandet2013domain}, prediction of a label $Y$ is based not just on a feature vector $X$ from a test domain, but also on the marginal distribution $P_{X|D}$ of feature vectors from the test domain 
% (which is known from the unlabeled test data)
. Thus, $P_{X|D}$ allows the predictor to adapt to the test domain.
% The first two works that formally proposed the DG problem are \citet{blanchard2011generalizing} and \citet{muandet2013domain}, where they learn predictions based not only on feature vectors, but also on domain-specific information. In their cases, the test $X$-marginal is used as additional input to the feature $X$ itself during inference time. 

Most subsequent work on DG, however, has sought to learn a classifier that predicts $Y$ solely from $X$, thus ignoring domain information during inference. This is reflected in
% in definitions 
recent surveys:
\begin{itemize}
% [leftmargin=10pt,align=left,labelwidth=\parindent,labelsep=4pt]
% [leftmargin=*]
    \item \citet{wang2022generalizing} (survey): ``The goal of domain generalization is to learn a robust and generalizable predictive function $h : \mathcal{X} \to \mathcal{Y} $ from the $M$ training domains to achieve a minimum prediction error on an unseen test domain $S_\text{test}$.''
    \item \citet{zhou2023domain} (survey): ``The goal of DG is to learn a predictive model $f : \mathcal{X} \to \mathcal{Y} $ using only source domain data such that the prediction error on an unseen
    % target 
    domain $T = \left\{ x^T \right\}$ is minimized.''
\end{itemize}

This focus on {\em domain agnostic} DG, in which domain information is not used at inference, can be traced largely to the datasets employed in DG research. While \citet{blanchard2011generalizing} studied an application in flow cytometry, most DG research since then has focused on applications in computer vision. 
For many vision tasks in the DG literature, there exists a single classifier that performs well on all domains. In such settings, strong performance is indeed possible without leveraging domain-specific information at test time.

% The focus on domain-agnostic classifiers means that 
As a consequence,
both the theory and methodology of DG remain relatively underdeveloped for settings in which the Bayes-optimal classifier varies substantially across domains. While some recent methods incorporate domain information into training and inference \citep{dubey2021adaptive, zhang2021adaptive, yao2024improving}, the fundamental role of domain information, namely, its precise decision-theoretic value and the conditions under which it strictly improves performance, remains poorly understood.

% \TODO{Yilun: double check the 3 citations}

To address this limitation, we introduce a theoretical framework for {\em domain-informed} DG. This framework is grounded in a novel statistical formulation of DG that extends both the original formulation introduced by \citet{blanchard2011generalizing} and the domain-agnostic setting. It allows that domain {\em metadata} --- features that describe the domain --- is available during training and inference. 

Leveraging this framework, we establish risk bounds that demonstrate the benefit of domain information under {\em posterior drift}, a new class of DG problems where the optimal classifier changes substantially with the domain (\Cref{thm:dg_improve,prop:dg_inf_lower_bound}). 
% Put another way, we show that under the posterior drift, domain-agnostic methods are provably inferior. 
In particular, we show that under posterior drift, domain-agnostic methods are provably suboptimal.
We further provide a condition, reflecting vision tasks as described above, under which domain-agnostic methods are sufficient.

To demonstrate the practical implications of our theory, we experimentally investigate the performance of domain-informed empirical risk minimization (DI-ERM), the natural ERM procedure induced by our formulation, and compare it to leading DG methods on language and vision tasks.

% Our contributions are summarized as follows:
% \begin{itemize}
%     \item A new statistical formulation of DG that incorporates domain metadata
%     \item Risk bounds that elucidate the decision-theoretic benefits of domain metadata, in particular under a proposed posterior drift class of DG problems
%     \item A formal condition under which domain-agnostic DG is sufficient and an explanation of the oft-repeated finding in the DG literature that domain-agnostic ERM is ``hard to beat'' 
%     % Yilun: seems to me this is a small point
%     \item A quantification of the 
%     % difference 
%     decision-theoretic gaps
%     between DG and domain adaptation, addressing an open question in \citet[][Lemma 9]{blanchard2021domain}.
%     % \item An explanation of the oft-repeated finding in the DG literature that domain-agnostic ERM is ``hard to beat,'' and a description of when this is actually the case
%     \item Empirical validation of our theory on both language and vision tasks.
% \end{itemize}

Our contributions are summarized as follows:
\begin{itemize}
% [leftmargin=\parindent,align=left,labelwidth=\parindent,labelsep=0pt]
% [leftmargin=10pt,align=left,labelwidth=\parindent,labelsep=4pt]
    \item A statistical formulation of DG that incorporates domain metadata and generalizes the standard DG setting.
    \item Risk bounds characterizing the decision-theoretic value of domain metadata, particularly under a newly identified \emph{posterior drift} class of DG problems.
    \item A formal condition under which domain-agnostic DG is sufficient, providing a theoretical explanation for the widely observed phenomenon that domain-agnostic ERM is ``hard to beat'' on standard benchmarks.
    \item A quantification of the decision-theoretic gap between DG and domain adaptation, resolving an open question posed in \citet[][Lemma~9]{blanchard2021domain}.
    \item Empirical validation of our theory on both language and vision benchmarks.
\end{itemize}

\section{Literature Review}

\citet{blanchard2011generalizing} introduced domain generalization (DG), motivated by a medical application involving the automatic gating of flow cytometry data. Since then, most DG research has focused on applications in computer vision. A typical DG task in this setting involves training models on labeled images from multiple visual domains (e.g., styles or rendering conditions) and evaluating generalization to a previously unseen domain. Benchmark datasets such as VLCS \citep{fang2013vlcs}, PACS \citep{li2017pacs}, OfficeHome \citep{venkateswara2017officehome}, DomainNet \citep{peng2019domainnet}, and ImageNet-Sketch \citep{wang2019imagenetsketch} have become standard in this line of work.

% These vision-based setups implicitly satisfies a \emph{universal Bayes classifier} (UBC) assumption:
In these vision-based setups, 
% the underlying distributional shift can be described as covariate shift with universal Bayes classifier (CSUBC): 
while the $X$-marginal changes across the domains, 
% as can the posterior, 
a single function of the input $X$ realizes the Bayes classifier for all domains.
% \citep{ben2006analysis, mansour2009domain}, where the marginal distribution \( P_X \) varies significantly across domains, 
% while the posterior $P_{Y|X}$ is constant.
Under this regime, domain-specific features are often regarded as irrelevant or even spurious \citep{Sagawa2020Distributionally, bai2025invariant} for predicting labels. Consequently, much of the literature has focused on learning \emph{domain-invariant} representations.
% that generalize robustly to unseen domains 
\citep{sun2016coral, ganin2016dann, arjovsky2019irm}. 
Additional references are in \Cref{appx:lit_review}.

Despite extensive efforts to design advanced DG algorithms, a persistent puzzle is the surprising effectiveness of what we will call pooling empirical risk minimization (pooling-ERM), a baseline that simply pools labeled data from all source domains together and trains a domain-agnostic classifier. Multiple studies have consistently found that pooling-ERM 
% remains highly competitive:
is hard to beat: 
% \citet{gulrajani2021search} report empirically that ``when carefully implemented and tuned, ERM outperforms the state-of-the-art 
%     in terms of average performance... 
%     no algorithm included in DomainBed (dataset) outperforms ERM by more than $1 \%$.''
% \citet{rosenfeld2021risks} show theoretically that ``Invariant Risk Minimization and its alternatives fundamentally do not improve over standard ERM.'' 
% Recent work by \citet{Teterwak2025WACV} further demonstrate empirically that ``the additional tuning in our improved baseline ERM++ outperforms both the prior ERM baselines and all recent SOTA methods on DomainBed.''

\begin{itemize}
% [leftmargin=10pt,align=left,labelwidth=\parindent,labelsep=4pt]
% [leftmargin=*]
    \item \citet{gulrajani2021search} (empirical): ``when carefully implemented and tuned, ERM outperforms the state-of-the-art 
    in terms of average performance... 
    no algorithm included in DomainBed (dataset) outperforms ERM by more than $1 \%$.''
    % \item \citet{koh2021wilds} (empirical): ``models trained with CORAL, IRM, and Group DRO generally fail to improve over models trained with ERM''
    \item \citet{rosenfeld2021risks} (theory): ``Invariant Risk Minimization and its alternatives fundamentally do not improve over standard ERM.'' 
    % \item \citet{gouk2024limitations} (theory): ``Our findings show that in all the DG settings we consider, it is not possible to significantly outperform ERM''
    \item \citet{Teterwak2025WACV} (empirical): ``the additional tuning in our improved baseline ERM++ outperforms both the prior ERM baselines and all recent SOTA methods on DomainBed.''
\end{itemize}
Similar observations have been made on other DG benchmarks \citep{koh2021wilds, sagawa2022wilds2}, as well as in related settings such as federated domain generalization \citep{bai2024benchmarking}. 

We argue that the key to understanding these observations lies in the probabilistic assumptions governing distributional shift across domains.
Different from prior works, our work examines an assumption called 
%is motivated by a class of problem characterized by 
\emph{posterior drift}, previously studied in domain adaptation \citep{scott2019generalized, cai2021transfer, maity2021linear, zhu2024label, wang2025transfer}, where the conditional distribution  of $Y | X$
%\( P_{Y|X} \) 
varies across domains. This type of shift commonly arises in natural language processing (NLP). For a given sentence \( X \), different annotators (or populations) may interpret its semantic content differently, leading to divergent labels \( Y \) (e.g., offensive vs. non-offensive, positive vs. negative). This form of ambiguity and annotator disagreement has been documented across a wide range of NLP applications \citep{de2019commitmentbank, plank-2022-problem, deng2023you}.

In this work, we formulate posterior drift in a DG context, and develop statistical theory for DG problems characterized by posterior drift.
Under this regime, pooling-ERM is provably suboptimal, and strictly stronger performance is achievable by domain-informed ERM (defined in the next section).
% , where domain-specific information is used both during training and at inference.
Our theory highlights the value of using domain metadata to perform domain adaptive prediction. Through this lens, \citet{yao2024improving} is closest to our work. They leverage domain metadata to re-weight prediction during inference time. 
% Our contribution lies in discovering posterior drift as the key factor and a theoretical framework that captures it.
%Our contribution lies in identifying posterior drift as the key structural property and in developing a unified decision-theoretic framework that characterizes its implications.

% This paper develops a theoretical framework for DG under posterior drift. 
% Under this regime, pooling-based ERM is provably suboptimal, and strictly improved performance is achievable by incorporating domain information at prediction time through \emph{domain-informed} ERM (defined in the next section). 
% From a practical perspective, the theory suggests leveraging domain metadata to perform domain-adaptive prediction. Viewed through this lens, the work of \citet{yao2024improving} is most closely related, as it uses domain metadata to reweight predictions at inference time. Our contribution lies in identifying posterior drift as the key structural property and in developing a unified decision-theoretic framework that characterizes its implications.

\section{
 A General 
% Probabilistic 
Statistical
Framework
 }

In standard classification, a random pair $(X, Y)$ is assumed to be drawn from a fixed joint distribution $P_{XY}$, where $X \in \mc{X}$ is a feature vector and $Y \in \mc{Y} = \{1,\ldots, K\}$ the corresponding class label\footnote{This section easily extends to regression, but subsequent sections are specific to classification.}. The goal is to learn a function $f: \mathcal{X} \to \mathcal{Y}$ that minimizes the risk
\[
% R(f) = 
\expect{(X, Y) \sim P_{XY}}{ \indicator{ 
% \left\{ 
f(X) \neq Y 
% \right\} 
} }.
\]
Domain generalization (DG) can be framed in a similar way. Let $\mathcal{D}$ denote a set of possible domains, where the term \emph{domain} is a synonym for a joint distribution of $X$ and $Y$. Let $D$ be a random variable on $\mathcal{D}$. Furthermore, let $M$ be a random variable on a space $\mathcal{M}$ that, intuitively, provides partial information about $D$. 
% The idea in DG is that 
In DG,
$D$ determines a distribution of $(X,Y)$, but is not observed. $M$ provides partial information (or \emph{metadata}) about $D$, and is thus useful at test time in adapting the classifier to the test domain. 
\begin{remark}
    While the availability of metadata $M$ will depend on the application, 
    one choice that is frequently viable is to take $M = P_{X|D}$, the marginal distribution $X$ for the given domain, which is known at test time through the unlabeled test sample. 
\end{remark}

Below we argue that the observability of $M$ is what makes DG distinct from standard classification in a certain decision-theoretic sense.
Formally, we assume that $(X,Y,M,D)$ are jointly distributed, with joint distribution denoted $P_{XYMD}$. This distribution induces several other distributions of interest in this paper. We follow convention in denoting marginal distributions by keeping the relevant subscripts. For example, $P_{XYD}$ denotes the joint distribution of $(X,Y,D)$ after $M$ is marginalized out. Similarly, $P_{XY}$ denotes the marginal distribution of $(X,Y)$.

For any fixed $d \in \mathcal{D}$, $P_{XY|D=d}$ is a joint distribution of $(X,Y)$. Note that our notation is somewhat redundant, as both $d$ and $P_{XY|D=d}$ refer to a domain, namely, a joint distribution of $(X, Y)$, but these two notations both play a role in our discussion.\footnote{\citet{blanchard2011generalizing, blanchard2021domain} use $P_{XY}$ to denote a random domain, whereas in our notation, a random domain is either $P_{XY|D}$ or just $D$. Our introduction of $D$ for a random domain allows us to use $P_{XY}$ for the ``average'' domain, which will be a critical concept in what follows.}

% For any fixed $d \in \mathcal{D}$, $P_{XY \mid D = d}$ denotes a joint distribution over $(X,Y)$. 
% Note that our notation is somewhat redundant, as both $d$ and $P_{XY \mid D = d}$ refer to a domain—namely, a joint distribution of $(X,Y)$—but each notation plays a distinct role in our discussion.\footnote{\citet{blanchard2011generalizing, blanchard2021domain} use $P_{XY}$ to denote a random domain. In contrast, in our notation a random domain is represented either by $D$ or by $P_{XY \mid D}$. Introducing the random variable $D$ allows us to reserve $P_{XY}$ for the ``average'' domain, a distinction that will be critical in what follows.}

To formalize the notion that $M$ is a partial summary of $D$, we assume that 
$(X,Y)$ and $M$ are conditionally independent, given $D$:
\begin{align}
\label{assump:m}
    % (X, Y) | D, M = (X, Y) | D
    P_{XY|D,M} = P_{XY|D}.
\end{align}

% P_{XY|D} = P_{XY|MD}$.
This implies that, given $D$, the joint distribution of $X$ and $Y$ does not change with knowledge of $M$. 
An important special case where this holds is when $M = g(D)$ for some deterministic $g: \mathcal{D} \to \mathcal{M}$. 
% A few illustrative examples are provided in \Cref{tab:domain-metadata}.
Table~\ref{tab:domain-metadata} lists several applications that illustrate this probabilistic framework, and will be empirically evaluated in \Cref{sec:experiments}.
%We illustrate this probabilistic framework with motivating examples in Table~\ref{tab:domain-metadata}, whose implications will be discussed throughout the paper.

% \twocolumn[
\begin{table*}[t]
% \vspace{-0.2cm}
\caption{Examples of domains and metadata in different tasks.}
% \vspace{-0.2cm}
\label{tab:domain-metadata}
% \small
% \renewcommand{\arraystretch}{1.5}
% \begin{tabular}{p{0.18\linewidth}p{0.15\linewidth}p{0.15\linewidth}p{0.20\linewidth}p{0.20\linewidth}}
% \toprule
% \multicolumn{1}{c}{\textbf{Task}} & \multicolumn{1}{c}{\textbf{Input} \(X\)} & \multicolumn{1}{c}{\textbf{Label} \(Y\)} & \multicolumn{1}{c}{\textbf{Domain} \(D\)} & \multicolumn{1}{c}{\textbf{Metadata} \(M\)} \\

\resizebox{1\linewidth}{!}{
\begin{tabular}{ccccc}
\toprule
\textbf{Task} & \textbf{Input} \(X\) & \textbf{Label} \(Y\) & \textbf{Domain} \(D\) & \textbf{Metadata} \(M\) \\
\midrule
\makecell{Sentiment analysis \\ (Multiple Annotators)} & Sentence & \makecell{Sentiment label \\ (e.g., positive, negative)} & \makecell{Annotator identity \\ (e.g., ``Annotator 1'')} & \makecell{Annotator's demographic profile \\ (e.g., age)}
% , which may correlate with subjective labeling tendencies 
\\
\midrule
\makecell{Review rating prediction \\ (Multiple Reviewers)} & Product review & \makecell{Numerical rating \\ (e.g., 1–5 stars)} & \makecell{Reviewer identity \\ (e.g., ``Reviewer 2'')} & \makecell{Unlabeled texts written by the reviewer \\ \( \{X_i\}_{i=1}^n \stackrel{iid}{\sim} P_{X|D=d} \)}
% , capturing writing and rating behavior 
\\
\midrule
\makecell{Image classification \\ across styles} & Image & \makecell{Object category label \\ (e.g., dog, car)} & \makecell{Image style \\ (e.g., photograph, painting)} & Textual description of style 
% (e.g., ``photograph'') used to inform style-aware classification 
\\
% \midrule
% Fraud detection & Transaction & Fraud/non-fraud & Credit cards & Transaction history of cards
% \\
\bottomrule
\end{tabular}
}
% \vspace{-0.2cm}
\end{table*}
% ]

The training data available to the learner is generated as follows: First, $N$ domains $d_1, \ldots, d_N$ are sampled iid from $P_D$, but not observed. Then, conditioned on these $d_i$, corresponding values $m_i$ are observed. In addition, for each $i$, $1 \le i \le N$, data $(x_{ij}, y_{ij})$ are sampled iid from $P_{XY|D = d_i}$, $1 \le j \le n_i$. In summary, the overall training data is
$$
\left( m_i, (x_{ij}, y_{ij})_{j=1}^{n_i} \right)_{i=1}^{N}.
$$

The goal of the learner is to produce a function 
% $f: \mathcal{M} \times \mathcal{X} \to \mathcal{Y}$ such that 
$f$ that accurately predicts labels on a new, random domain. In particular, $f$ should minimize the risk
$$
R(f) :=
    % \expect{X, Y, M, D}{ \indicator{f(M,X) \neq Y} }.
    \expect{X, Y, M, D}{ \indicator{f(\cdot) \neq Y} }.
$$
% \vspace{-1em}
where the argument of $f(\cdot)$ depends on settings.
In practice, this risk is estimated by holding out several of the domains, and averaging the test errors on them.

\begin{remark}
This probabilistic framing of DG 
% is closely aligned to 
generalizes
the original formulation of DG by \citet{blanchard2011generalizing}.
% that of \citet{blanchard2011generalizing, blanchard2021domain}.
They focus on the special case where $M$ is the marginal distribution of $X$ for the given domain ($M = P_{X|D}$), and focus on the challenges associated with learning from empirical samples of the training and testing $X$-marginals.
\end{remark}

The training setup described above naturally gives rise to two different ways of using the available data. On one hand, the learner may choose to ignore the domain information and simply pool together all training samples, treating them as if they were drawn iid from a single domain. On the other hand, the learner may choose to leverage the observed metadata $m_i$, which serves as side information about the underlying domain. These two strategies lead to two corresponding empirical risk minimization principles.
% \paragraph{
% \textbf{Two ERMs}
% }
% $\quad$
Thus, let \( \mathcal{F} \subset \{ \mathcal{X} \times \mathcal{M} \to \mathcal{Y} \} \) denote a class of functions that take both instance $x$ and metadata $m$ as input, and \( \mathcal{G} \subset \{ \mathcal{X} \to \mathcal{Y} \}  \) a class of functions that take only $x$ as input.
Consider two empirical risk minimizers:\\
\textbf{Pooling-ERM:}
\begin{align}
\fxhat = \argmin{f \in \mathcal{G}} \frac1{N} \sum_{i=1}^N \frac1{n_i} \sum_{j=1}^{n_i} \ell(y_{ij},f(x_{ij})). \label{eqn:pooling_erm}
\end{align}
\textbf{Domain-informed (DI) ERM:}
\begin{align}
\fxihat = \argmin{f \in \mathcal{F}} \frac1{N} \sum_{i=1}^N \frac1{n_i} \sum_{j=1}^{n_i} \ell(y_{ij},f(x_{ij}, m_i)). \label{eqn:augmented_erm}
\end{align}
To highlight the importance of using domain metadata, we are interested in when DI-ERM outperforms pooling-ERM. 
In the next section, we compare the optimal risks associated with these two approaches. 
% to DG. 

% we work in the large-sample and “large-model” limit (where $\mathcal{F}$ and $\mathcal{G}$ can approximate the Bayes-optimal predictor arbitrarily well). 
% In this regime, standard learning-theoretic arguments imply that the performance of the two approaches is characterized by their corresponding Bayes risks, defined below.

% \newpage

\section{Risk and Bayes Risk 
% in domain generalization
}

To aid in understanding domain generalization, it is helpful to understand the risk and Bayes risk depending on the information available to the classifier. In this section, we examine the risk 
% prediction problems. 
%These settings differ only in what information the classifier $f$ has access to. In all cases, the performance measure is the risk
\begin{align*}
    R(f) :=
    \expect{X, Y, M, D}{ \indicator{f(\cdot) \neq Y} },
\end{align*}
where the argument of $f(\cdot)$ is determined by one of three possible settings. 
%depends on the setting. In this section, we compare the risks and Bayes risks associated to these three settings. 
This sets the stage for our main results in the next section.

\textbf{No Domain Information:} 
$\quad$
In this setting, the classifier only has access to the feature vector $x$ at test time, and is thus $f(x)$. As noted earlier, most empirical DG methods, especially in computer vision, have this form. The risk in this case is
    \begin{align*}
        R(f) & = \expect{X,Y,M,D}{ \indicator{f(X) \neq Y} } \\
        & = \expect{X,Y}{ \indicator{f(X) \neq Y} } \\
        & = \expect{X}{\expect{Y|X}{ \indicator{f(X) \neq Y} } },
    \end{align*}
where, because $f$ does not depend on $D$ or $M$, these variables marginalize out. Therefore, the problem reduces to learning with respect to the marginal distribution of $(X,Y)$, which can be viewed as the ``average'' domain. %corresponds to pooling data across domains. 
The optimal classifier $f^*_{\mathrm{pool}}$ is thus the Bayes classifier for the marginal distribution of $(X,Y)$:
\begin{align*}
    % f^*_{\mathrm{pool}}(x)
     \fx(x) = \argmax{k} \prob{Y=k|X=x}. 
    % \\
    % \text{some proposal: } \fx(x), f^*_{\mathrm{full}}(x,d), 
    % % f^*_{\mathrm{meta}}(x,m), 
    % f^*_{\mathrm{DI}}(x, m); 
    % f^*(x), f^*_{D}(x,d), f^*_{M}(x,m)
\end{align*}
The corresponding Bayes risk, $\rx$, is the Bayes risk for the marginal distribution of $(X,Y)$: 
    \begin{align*}
        % R^*(X)
        \rx
        := & \ \expect{X, Y}{ \indicator{\fx(X)\neq Y} } \\
        = & \ \expect{X}{ 1 - \max_{k} \prob{Y=k|X} }.
            % \\
            % & = \expect{D}{ \expect{X|D}{ 1 - \max_k \prob{Y=k|X} } }
    \end{align*}
$\rx$ is the best possible performance of pooling-ERM.
% This is the setting implicitly assumed by the majority of DG benchmarks.
% The limitation of it lies in that it does not take into account any test-time signal that can potentially lead to refined prediction - which we will analyze carefully in the latter section.

% \paragraph{
\textbf{Partial Domain Information:} %$f(X, M)$
% }
%The original formulation of DG by ~\citet{blanchard2011generalizing} assumed access to partial domain information $M = P_{X|D=d}$. Here we consider a more general setup of this, $M = g(D)$ for some $g$. 
$\quad$
This is our setting of domain generalization. The classifier has access to not only $x$, but also the metadata $m$ that conveys partial information about the true domain $d$. A classifier in this setting is denoted $f(x,m)$. The risk is
\begin{align*}
        R(f) &= \expect{X,Y,M,D}{ \indicator{f(X,M) \neq Y} } \\
        & = \expect{X,Y,M}{ \indicator{f(X,M) \neq Y} } \\
        & = \expect{X,M}{\expect{Y|X,M}{ \indicator{f(X,M) \neq Y} } }.
    \end{align*}
The optimal ``domain-informed'' classifier $\fxi$ is now the Bayes classifier for the distribution of $X,Y|M$,
\begin{equation*}
    \fxi(x,m) = \argmax{k} \prob{Y=k|X=x,M=m},
\end{equation*}
and the corresponding Bayes risk is
% \TODO{DG to DI}
\begin{align*}
    % R^*(X,M) 
    \rxi
    := & \ \expect{X, Y, M}{ \indicator{\fxi(X, M) \neq Y} }  \\
      = & \ \expect{X, M}{1 - \max_{k} \prob{Y=k|X,M} }. 
\end{align*}
$\rxi$ is the best possible performance of DI-ERM. Clearly, $\rx \ge \rxi$, since adding features (here, the metadata) can only decrease the Bayes risk.
% In this 
% % more informative 
% setting, 
% the optimal classifier uses both $X$ and domain-specific signal $M$ to predict $Y$.
% \begin{align*}
%     f^*(X, M) & = \argmin{f} \ \expect{(X, Y, D, M)}{ \indicator{f(X, M) \neq Y} } \\
%     & = \argmax{k} \prob{Y=k|X,M} \\
% \end{align*}
% This setting aligns with the original theoretical motivations of DG and highlights the value of leveraging test-time domain information. A key goal of our work is to reassert the importance of this formulation and demonstrate both its theoretical advantages and empirical benefits, particularly in contrast to the more commonly used $f(x)$ setting.

% \paragraph{
\textbf{Full Domain Information:} %$f(X, D)$
% }
$\quad$
In this setting, the classifier has full knowledge of the domain $D$ at test time, and is thus denoted $f(x,d)$. In practice, full knowledge of $D$ is not available, and this setting therefore serves as a bound on the best possible performance of DG. The risk in this setting is
\begin{align*}
        R(f) & = \expect{X,Y,M,D}{ \indicator{f(X,D) \neq Y} } \\
        & = \expect{X,Y,D}{ \indicator{f(X,D) \neq Y} } \\
        & = \expect{X,D}{\expect{Y|X,D}{ \indicator{f(X,D) \neq Y} } }.
    \end{align*}
The optimal classifier $\fxd$ is now the Bayes classifier for the distribution of $X,Y|D$,
   \begin{equation*}
        \fxd(x,d) = \argmax{k} \prob{Y=k|X=x,D=d},
    \end{equation*}
and the corresponding Bayes risk is: 
\begin{align*}
        % \fxd(x,d)& = \argmin{f} \ \expect{D}{ \expect{X, Y|D}{ \indicator{f(X, D) \neq Y} } } \\
        %     % & = \argmin{f} \ \expect{(X, Y, D)}{ \indicator{f(X, D) \neq Y} } \\
        %     % & = \argmin{f} \ \expect{(X, D)}{ \expect{Y|X,D}{ \indicator{f(X, D) \neq Y} } } \\
        %     & = \argmax{k} \prob{Y=k|X,D}. \\
        % R^*(X, D) 
        \rxd
        := & \ \expect{X, Y, D}{ \indicator{\fxd(X,D)\neq Y} } \\
         = & \ \expect{X, D}{ 1 - \max_k \prob{Y=k|X,D} }. 
\end{align*}
% $R^*(X,D)$ 
$\rxd$ serves as a lower bound for the risk in domain generalization, as described in the next section.

% In this setting, the classifier has full knowledge of the test domain, in other words, the joint distribution of $(X,Y)$ for the given test domain. Therefore, $R^*(X,D)$ is the Bayes risk for the test domain, which serves as a lower bound for the risk in domain generalization. 

%This setting serves as an oracle and is typically infeasible in DG, as the domain identity $D$ is unobserved at test time. In practice, this corresponds to the absence of labeled data from the target domain.

\begin{remark}
\label{remark:dg_da}
DG is very similar to multi-source domain adaptation (DA). The training data available to the learner is the same in both problems. In DA, however, the performance metric of interest is 
% $R^*(X,D)$
$\rxd$. 
Essentially, the difference between DG and DA is that DG seeks to optimize the expected risk on a new, random test domain, whereas DA seeks to optimize risk on a specific, fixed test domain. 
% To achieve this goal, it is common in DA to make additional assumptions about how the test domain relates to the training/source domains. In DG, no such assumptions are necessary since the Bayes classifier only uses observable information. 
\end{remark}

\section{Risk Comparison}

This section develops bounds that relate the three Bayes risks defined in the previous section, 
% revealing settings where domain information is and is not beneficial. 
identifying when domain information is beneficial and when it is not.
The following basic result provides a starting point.
%We begin with a foundational principle guiding this work: conditioning on domain-specific information—whether complete or partial—cannot degrade classification performance. This principle yields the following risk hierarchy:
%The three Bayes risks are related as follows.
\begin{proposition}[Risk Hierarchy]
% \[
  $$
  \rx \geq \rxi \geq \rxd.
  $$
% \]
\label{prop:risk_hierarchy}
\end{proposition}
\vspace{-1.4em}
The proof is straightforward (see \Cref{subsec:risk_hierarchy_proof}). 
The first inequality 
% is trivial and was noted previously.
holds as adding more feature can never decrease the Bayes risk. 
The second inequality follows from Assumption \eqref{assump:m}.

Our goal in this section is to describe distributional assumptions under which these 
%We next describe distributional assumptions under which these 
inequalities become strict, and with a quantifiable gap. 
%Under such assumptions, domain information yields improved predictive capability compared to prediction with feature vectors only. 
Toward that end, consider the following. % definition.
\begin{definition}[Point-wise Margin]
Consider any random triple $(X, Y, M)$, where $Y$ is discrete.
Define the point-wise margin of $Y|X=x, M=m$ as,
\begin{align*}
    % \forall m \sim P_M, \ 
    \gamma (x, m) := & \max_k \prob{Y=k|X=x, M=m} \\
    & - \operatorname{2nd} \max_k \prob{Y=k|X=x, M=m}.
\end{align*}

\end{definition}
The operator $\operatorname{2nd} \max_k$ returns the second largest value of its argument. Thus, if the two largest values of $\prob{Y=k|X=x, M=m}$ are the same, $\gamma(x,m) = 0$.
Intuitively, $\gamma(x,m)$ reflects the degree of certainty that the Bayes classifier $f^*_{\mathrm{DI}}(x,m)$ has about its prediction. The larger $\gamma(x,m)$, the more confident the prediction.
%\TODO{give a precise definition; think about tie-breaking}

The next result gives upper and lower bounds on the gap between $\rxi$ and $\rx$. This gap is the additional reduction in risk that results from leveraging the partial domain information $M$.

% \newpage

\begin{figure}[!htbp]
    \centering
    \includegraphics[width=\linewidth]{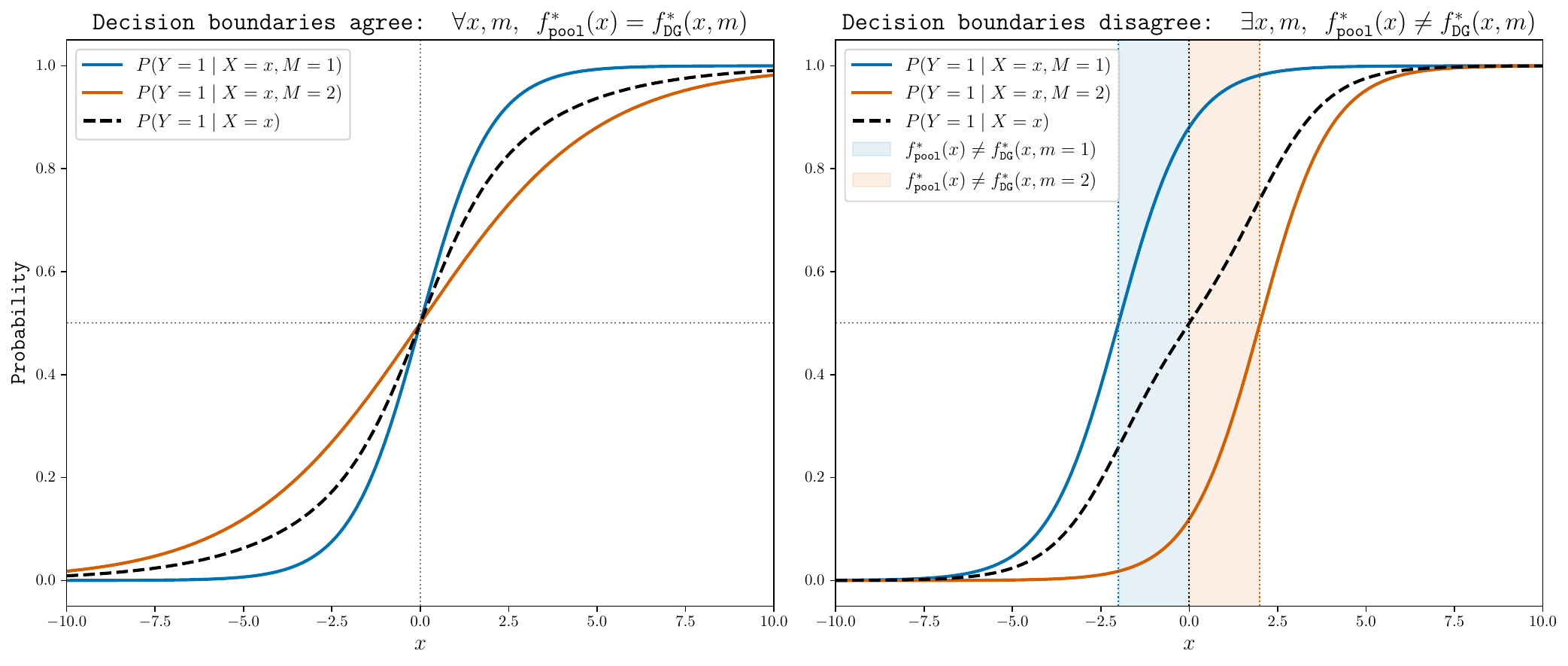}
    \caption{Illustration of \Cref{thm:dg_improve}. Consider binary classification with $X \in \mathbb{R}$, $Y \in \{1, 2\}$, and $M \in \{1,2\}$. Then the Bayes classifiers $\fx(x)$, $\fxi(x, m=1)$ and $\fxi(x, m=2)$ can be obtained by thresholding the corresponding posteriors at $1/2$. The left figure shows a scenario where the domain-informed classifier \( \fxi \) and the pooled classifier \( \fx \) agree everywhere, and therefore both upper and lower bound are $0$. In this case, domain information $M$ is not beneficial. The right figure shows a scenario where \( \fxi \) disagrees with \( \fx \) in certain regions, and domain information does lead to lower Bayes risk. }
    \label{fig:risk_reduction}
\end{figure}

\begin{theorem}[Risk Reduction w/ Domain Info] Consider any random triple $(X, Y, M)$, where $Y$ is discrete.
Then
\begin{align*}
    & \expect{X, M}{ \gamma(X, M) \indicator{  \fx(X)\neq \fxi(X, M) } } \\
    & \quad \leq \rx - \rxi \\
    & \quad \leq \expect{X, M}{ \indicator{ \fx(X)\neq \fxi(X, M) } }.
\end{align*}
% Furthermore, equality on both sides holds if and only if \TODO{seems to be only ``if''}
% \begin{align*}
%     \forall x, m, \ \argmax{k} \prob{Y=k | X = x} = \argmax{k} \prob{Y=k | X=x, M = m}.
% \end{align*}

\label{thm:dg_improve}
    
\end{theorem}
The proof of Theorem \ref{thm:dg_improve} is in \Cref{subsec:dg_improve_proof}. 
% The proof proceeds by first proving the inequality conditioned on $X$, and then averaging over $X$. 
% To gain some intuition, we illustrate this ``conditioned on $X$'' form of the result with the following example. 
The upper bound represents the probability of disagreement between the domain-informed classifier \( \fxi \) and the pooling classifier \( \fx \). The lower bound can be interpreted as the expected cost of disagreement, where the cost is zero when the predictions agree, and equals to the margin \( \gamma(X, M) \) when they differ. Hence, domain information is particularly beneficial when \( \fxi \) frequently disagrees with \( \fx \) in regions of high confidence. 
\Cref{thm:dg_improve} also implies necessary and sufficient conditions for the risk gap being zero. If $\fxi = \fx$ almost surely, then $\rx - \rxi = 0$. If $\rx - \rxi = 0$, then $\gamma(X, M) \indicator{  \fx(X)\neq \fxi(X, M) }$ must be zero almost surely, meaning $\fxi$ and $\fx$ can only disagree whenever two or more classes achieve the maximum posterior probability.
\Cref{fig:risk_reduction} gives more intuition.

\subsection{Decision-Theoretic Gain Under Posterior Drift}
\label{subsec:pd_gain}

%Theorem \ref{thm:dg_improve} holds regardless of the distribution $P_{XYMD}$. 
Theorem \ref{thm:dg_improve} holds regardless of the distribution $P_{XYMD}$. 
By imposing structural assumptions, stronger conclusions can be drawn.
We now examine a class of distributions where the gap $\rx - \rxi$ has a more concrete lower bound.
This class of distributions is motivated by applications, particularly in NLP, where the posterior \( P_{Y \mid X, D} \) (correspondingly, $P_{Y \mid X, M}$) differs across domains due to inherent ambiguity or subjectivity. 
A canonical example is the sentiment or toxicity annotation task, where annotators often disagree on the same text.
For instance, in the age-related sentiment analysis dataset of \citet{diaz2018addressing}, the sentence
\textit{``Old people's appearance contains so much lived life.''}
received conflicting labels: 2/5 annotators seeing it as `very positive', 2/5 as `somewhat positive', and 1/5 as `very negative'. 
This reflects how labeling tendency varies with annotator identity (aka, domain).
We capture this phenomenon by introducing a formal posterior drift class for DG.

\begin{definition}[Posterior Drift for DG] For $\gamma, \epsilon > 0$, denote
% \resizebox{\linewidth}{!}{%
% \begin{minipage}{\linewidth}
\begin{align*}
    \Pi(\gamma, \epsilon) := \Big\{
        & P_{XYMD}:  \forall x, m, \ \gamma(x, m) \geq \gamma, \text{ and } \\
        & P_{XMM'} \Big( \fxi \big(X, M \big) 
        % \\
        % & \quad 
        \neq \fxi \big(X, M' \big) \Big) \geq \epsilon
    \Big\},
\end{align*}
% \end{minipage}%
% }
% }
where $(M, M') \mid X \sim P_{M|X} \otimes P_{M|X}$ are two independent draws.

\label{def:pd_dg}

\end{definition}

This class of DG problems captures settings where optimal classifiers with different $M$ make conflicting predictions on a non-negligible region of the input space. 
The parameter $\gamma$ quantifies the point-wise confidence of the optimal predictor, and
the parameter \( \epsilon \) quantifies the average amount of variation in $P_{Y|X, M}$ for different $M$.
With this, we have an explicit lower bound:

\begin{theorem}[Gain Under Posterior Drift]
    \label{prop:dg_inf_lower_bound}
    \begin{align*}
        \inf_{P_{XYMD} \in \Pi(\gamma, \epsilon)} \Big[  
      % R ^*(X) - R^*(X, M) 
      \rx - \rxi
      \Big] \geq  \frac{\gamma \cdot \epsilon }{2}.
    \end{align*}
\end{theorem}
% \vspace{-2em}
% The proof is in \Cref{subsec:dg_inf_lower_bound_proof}. This lower bound shows that leveraging domain-specific information yields a provable benefit of at least $\gamma \epsilon /2$ for this particular formulation of posterior drift. It is also a more concrete case of the lower bound in \Cref{thm:dg_improve}.
The proof is given in \Cref{subsec:dg_inf_lower_bound_proof}.  
This result establishes that, under the posterior drift class $\Pi(\gamma,\epsilon)$, the benefit of leveraging domain-specific information is guaranteed to be at least $\gamma \epsilon / 2$.  
%In other words, the risk reduction from domain-informed prediction is not only possible but provably guaranteed in this regime.  
This result can be viewed as a concrete version of the more general lower bound in \Cref{thm:dg_improve}.

% Yilun: I am relating to the conclusion in our label noise paper
\begin{remark}
    In contrast to pessimistic results in \emph{domain adaptation}, where no method consistently outperforms vanilla ERM under posterior drift \citep{zhu2024label, wang2025rethinkingdistributionshiftsempirical}, our work presents an optimistic view in \emph{domain generalization}: by conditioning on domain metadata \( M \), we can provably do better than pooling-based prediction.
\end{remark}

\subsection{DG vs. Full Domain Knowledge}

% Although not the focus of this paper, 
A version of 
\Cref{thm:dg_improve} also holds for the gap \( \rxi - \rxd \), where $\rxd$ is the risk of a classifier that has full knowledge of the test domain.\footnote{$\rxd$ is the performance measure of interest in multi-source domain adaptation, see \Cref{remark:dg_da}.} Such a bound addresses a question left open by \citet[][Lemma 9]{blanchard2021domain}, who established that this gap is lower bounded by zero, and provide a condition under which the gap equals zero. The following result bounds this gap in a more general setting.
\begin{theorem} [$\rxi$ vs. $\rxd$]
    \label{thm:full_dg_gap}
    Let 
    \begin{align*}
        \widetilde{\gamma}(x,d) := & \ \max_k \prob{Y=k|X=x, D=d} \\
        & \quad - \operatorname{2nd} \max_k \prob{Y=k|X=x, D=d}.
    \end{align*}
    Then 
    \begin{align*}
        & \expect{X, D, M}{ \widetilde{\gamma}(X, D) \indicator{  \fxd(X,D)\neq \fxi(X, M) } } \\
        & \quad \leq \rxi - \rxd \\
        & \quad \leq \expect{X, D, M}{ \indicator{ \fxd(X,D)\neq \fxi(X, M) } }.
    \end{align*}
    % and in particular, 
    % \begin{align*}
    %     & \fxd(X,D) = f^*_{\mathrm{DI}}(X,M) \\
    %     & \quad \text{ almost surely w.r.t. } P_{XMD} \\
    %     & \implies \rxd = \rxi.
    % \end{align*}
\end{theorem}
% The result has an interpretation analogous to that of \Cref{thm:dg_improve}. In particular, if the observed domain information is of low quality, in the sense that $\fxd$ disagrees with $\fxi$ often, and with high confidence, then $\rxi$ can be substantially worse than $\rxd$. 
% Necessary and sufficient conditions for this gap to be zero are also analogous to those for \Cref{thm:dg_improve}.

The result parallels \Cref{thm:dg_improve}: if domain information is low quality, i.e., $\fxd$ often disagrees with $\fxi$ with high-confidence, then $\rxi$ can be much larger than $\rxd$. The necessary and sufficient conditions for this gap to vanish are analogous to those in \Cref{thm:dg_improve}.

% \begin{remark}
%     Although not the focus of this paper, a version of 
% \Cref{thm:dg_improve} also holds for the gap \( \rxi - \rxd \), where $\rxd$ is the risk of a classifier that has full knowledge of the test domain. This bound quantifies the difference between DG and domain adaptation, and addresses a question left open by \citet[][Lemma 9]{blanchard2021domain}, see \Cref{appx:partial_vs_full} for more detailed discussion.
% % who established that this gap is lower bounded by zero, and provide a condition under which the gap equals zero. The following result bounds this gap in a more general setting.
% \end{remark}

\subsection{Gain Under UBC?}
\label{sec:di_erm_cov_shift}
The preceding subsection establishes that domain information can strictly improve performance under posterior drift.
% , where the Bayes-optimal classifier varies across domains. 
We now contrast this with the \emph{universal Bayes classifier} (UBC) regime, in which a single function of the input $X$ realizes the Bayes classifier for all domains.
% , even though the marginal distribution of $X$ may vary.
\begin{definition}[Universal Bayes Classifier]
    A DG problem satisfies \emph{universal Bayes classifier} assumption if
    \begin{align*}
        \fx(x) = \fxm(x, m) = \fxd(x, d) \quad \forall x, m, d.
    \end{align*}
\end{definition}
This is inspired by applications in vision where a single classifier performs well across all domains.
A concrete example of this is when the support of $P_{X|D}$ is disjoint across domains.
The next result follows from \cref{thm:dg_improve}.
% Covariate shift is a classical example of such a setting: as the domain information $M$ varies, the marginal distribution $P_{X \mid M}$ may change, while the conditional distribution $P_{Y \mid X, M}$ remains fixed \citep{candela2008dataset}. 
% More generally, we refer to any DG problem satisfying the \emph{universal Bayes classifier} (UBC) assumption, under which the Bayes classifier $\fxi$ does not depend on $M$.

% In this regime, access to domain-specific information cannot reduce the Bayes risk.

% Theorem \ref{thm:dg_improve} holds regardless of the distribution $P_{XYMD}$. By considering assumptions on this distribution, stronger conclusions may be drawn.
% Covariate shift refers to a setting where, as the domain information $M$ varies, $P_{X|M}$ changes, but $P_{Y|X,M}$ does not \citep{candela2008dataset}. 
% More generally, we extend the meaning of covariate shift into universal Bayes classifier
% (UBC) assumption, aka, any DG problem where the Bayes classifier $\fxi$ does not depend on $M$. 

\begin{corollary}  
    Under UBC, $\rx = \rxi$.
    \label{cor:cov_shift}
\end{corollary}

% Still, empirical works has shown performance gain of leveraging domain information even under covariate shift \citep{dubey2021adaptive, bui2021exploiting}. Our next statement shows that this is an effect of function class.
% This is a result of \Cref{thm:dg_improve}, which shows domain metadata do not exhibit decision-theoretic benefit.

In such a scenario, access to domain-specific information cannot reduce the Bayes risk. 
Nevertheless, studies have reported empirical performance gains from incorporating domain information even under UBC \citep{dubey2021adaptive, bui2021exploiting}. 
% Our next result shows that such gains arise not from information-theoretic advantages, but from the choice of function class.  
Our next result shows that such gains 
% is not a result of decision-theoretic benefit, but instead 
arise from the choice of function classes.

% \TODO{may need to address bui2021 better, there's an aistats reviewer who holds strong opinion on that}

% \begin{definition}[DI and Pooling Class]
%     \label{def:di_pool_class}
%     Define a sequence of domain-informed function classes $\mathcal{F}_1, \mathcal{F}_2, \ldots \subset \{ \mathcal{X} \times \mathcal{M} \to \mathcal{Y} \}$ and corresponding pooling function classes $\mathcal{G}_1, \mathcal{G}_2, \ldots \subset \{\mathcal{X} \to \mathcal{Y} \}$ as follows:
%     \begin{enumerate}
%         \item  $\forall g \in \mathcal{G}_i$, $\exists f \in \mathcal{F}_i$ such that $g(x) = f(x, m_0)$ for some fixed $m_0 \in \mathcal{M}$.
%         \item $\mathcal{G}_1, \mathcal{G}_2, \ldots$ satisfies universal approximation property, i.e., for all $P_{XY}$, $\inf_{g \in \mathcal{G}_k} R(g) \to \rx$ as $k \to \infty$.
%     \end{enumerate}
% \end{definition}

\begin{assumption}[DI \& Pooling Function Classes]
\label{ass:di_pool_class}
Let $\{\mathcal{F}_k\}_{k=1}^\infty \subset \{ \mathcal{X} \times \mathcal{M} \to \mathcal{Y} \}$ and  
$\{\mathcal{G}_k\}_{k=1}^\infty \subset \{ \mathcal{X} \to \mathcal{Y} \}$ be sequences of function classes related as follows:
\begin{enumerate}
    \item For every $g \in \mathcal{G}_k$, there exists $f \in \mathcal{F}_k$  such that  
    \[
        g(x) = f(x, m) \quad \forall m \in \mc{M}.
    \]
    \item The pooling classes $\{\mathcal{G}_k\}$ satisfy the universal approximation property: for any distribution $P_{XY}$,  
    \[
        \inf_{g \in \mathcal{G}_k} R(g) \;\to\; \rx \quad \text{as } k \to \infty.
    \]
\end{enumerate}
\end{assumption}

% The first condition is satisfied by common architectures (e.g., feedforward ReLU networks) and $\mathcal{G}$ can be induced from $\mathcal{F}$ by setting the weights from the $m$ inputs to zero. What's more, $\mathcal{G}$ satisfies the universal approximation property as the network size grows.

The first condition requires that the pooling class $\mathcal{G}_k$ can be represented by the domain-informed class $\mathcal{F}_k$. This condition is satisfied by common architectures such as feedforward ReLU networks, where $\mathcal{G}_k$ can be induced from $\mathcal{F}_k$ by setting the weights associated with the metadata inputs to zero.
The second condition assumes that the pooling classes $\{\mathcal{G}_k\}$ are sufficiently expressive to approximate $\fx$.

\begin{proposition}
% [Risk Hierarchy Under Covariate Shift]
\label{prop:risk_hierarchy_class}
For function classes $\mathcal{F}_k$ and $\mathcal{G}_k$ satisfying Assumption~\ref{ass:di_pool_class}, denote 
\begin{align*}
    R^*_{\mathrm{pool},\mathcal{G}_k} := \inf_{g \in \mathcal{G}_k} R(g) \ \text{ and } \ R^*_{\mathrm{DI},\mathcal{F}_k} := \inf_{f \in \mathcal{F}_k} R(f).
\end{align*}

Under UBC, 
% the following 
% %risk hierarchy 
% holds:
\begin{equation}
    % R_{\mathrm{pool},\mathcal{G}_k} := \inf_{g \in \mathcal{G}_k} R(g)  \geq 
    % R_{\mathrm{DI},\mathcal{F}_k} := \inf_{f \in \mathcal{F}_k} R(f), 
    R^*_{\mathrm{pool},\mathcal{G}_k} \geq R^*_{\mathrm{DI},\mathcal{F}_k} \quad \forall k, 
    \label{eqn:cs_asymptotic}
\end{equation}
and
\begin{equation}
    \lim_{k \to \infty} R^*_{\mathrm{pool},\mathcal{G}_k}  = \lim_{k \to \infty} R^*_{\mathrm{DI},\mathcal{F}_k}.
% \end{align*}
\end{equation}

\end{proposition}

% \begin{proof} (will be put into appendix)
%     The first line follows from the first assumption of of Assumption~\ref{ass:di_pool_class}, where every function $g \in \mathcal{G}_k$ is also realizable in $\mathcal{F}_k$.
%     The second line follows from covariate shift assumption and the universal approximation property of function classes.
% \end{proof}

This shows that with a restricted function class $\mathcal{G}_k$, leveraging domain-information may lead to a benefit under UBC. Indeed, strict inequality in \Cref{eqn:cs_asymptotic} is possible, as illustrated with a simple example in \Cref{sec:cs_eg}. However, these gains vanish asymptotically as the function classes become more expressive. This phenomenon is also observed empirically in \Cref{sec:experiments}.

\section{Experiments}
\label{sec:experiments}

We evaluate the effectiveness of domain-informed ERM (DI-ERM) in three experimental settings. 
The goal is not to achieve state-of-the-art performance, but to empirically validate the theoretical insights from previous sections regarding the benefits of leveraging domain metadata under posterior drift (e.g., \Cref{thm:dg_improve,prop:dg_inf_lower_bound}).

\paragraph{Methods} 

% Our primary focus is on the comparison between DI-ERM and pooling-ERM.
% Other baselines like IRM \citep{arjovsky2019irm}, GroupDRO \citep{Sagawa2020Distributionally}, and CORAL \citep{sun2016coral} are popular DG methods that leverage domain index during training time, but predict only using $X$ during inference time.
% Closer to our setup, D3G \citep{yao2024improving} leverages domain metadata during both training and testing. 
% They train a separate prediction head for each domain and use metadata to re-weight predictions during inference time. As we will see, D3G struggles when the number of domains is large.

Our primary comparison is between DI-ERM and pooling-ERM. 
This comparison directly tests our central theoretical claim: incorporating domain metadata at inference time can strictly improve performance under posterior drift, while pooling-ERM is suboptimal.

We also include several widely used DG baselines: IRM \citep{arjovsky2019irm}, GroupDRO \citep{Sagawa2020Distributionally}, and CORAL \citep{sun2016coral}. 
These methods exploit domain indices during training but make predictions using only the input $X$ at inference time, and therefore fall within the domain-agnostic paradigm.

Closest to our setting, D3G \citep{yao2024improving} also leverages domain metadata at both training and inference time. 
Unlike DI-ERM, D3G trains domain-specific prediction heads---one for each training domain---and uses metadata to reweight their predictions during inference. 
This design can become less robust when the number of training domains is large relative to the amount of labeled data per domain. 
In contrast, DI-ERM encodes domain metadata directly as additional input, yielding a simpler and empirically more effective approach in such regimes.

% Closest to our setting, D3G \citep{yao2024improving} also leverages domain metadata at both training and test time. 
% Unlike DI-ERM, D3G trains domain-specific prediction heads (for each training domain) and uses metadata to reweight predictions during inference, 
% which indicates this strategy may become less robust when the number of training domains is large while labeled training data per domain is small. 
% DI-ERM encode domain metadata $m$ as input. It is simpler yet more effective.
% , highlighting a key practical distinction between DI-ERM and alternative domain-informed approaches.

Additional results, including linear probing, benchmarks against alternative and state-of-the-art methods, and complete experimental details, are provided in \Cref{appx: experimental-details}.

\subsection{Annotator Disagreement
%Sentiment Disagreement Among Annotators
 }
\label{subsec:exp_annotator}
% \paragraph{Sentiment disagreement among annotators}
In many NLP tasks, annotators exhibit subjective preferences, leading to disagreement on the label $y$ for the same input $x$—a form of posterior drift discussed in \Cref{subsec:pd_gain}. 
To study this phenomenon, we use the dataset of \citet{diaz2018addressing}, which re-annotates a subset of the Sentiment140 dataset \citep{go2009twitter} for training and provides a test set drawn from age-related blog posts.
The training set comprises 59{,}235 sentences labeled by 1{,}481 annotators; the test set includes 1{,}419 sentences labeled by 878 annotators. 
Each sentence is annotated by 4–5 individuals, and the labels exhibit high disagreement (about 40 $\%$), a sign of large posterior drift. In this setting, the input $x$ is a sentence, the label $y \in \{1,2,3,4,5\}$ denotes sentiment on a five-point scale, the domain $d$ corresponds to an annotator, and the domain information $m$ consists of demographic metadata (e.g., age, upbringing region).

% To encode domain information $M$, we concatenate it with the sentence $x$ in a text-prompt format, as illustrated in \Cref{fig:annotator_prompt}. We finetune the BERT base model \citep{devlin2019bert}. \Cref{tab:dierm_annotator} reports the results. DI-ERM substantially outperforms pooling-ERM, demonstrating that leveraging annotator metadata can dramatically improve predictive accuracy. Notably, DI-ERM also surpasses the previous state-of-the-art reported by \citet{deng2023you}.

To encode domain information $M$, we concatenate it with the sentence $x$ in a text-prompt format (\Cref{fig:annotator_prompt} in \Cref{appx: experimental-details}). We finetune the BERT base model \citep{devlin2019bert}. \Cref{tab:dierm_annotator} reports the results. Since prior DG methods have not been evaluated on this dataset, we implemented and benchmarked them under the same experimental setup. DI-ERM substantially outperforms pooling-ERM, demonstrating that leveraging annotator metadata can dramatically improve predictive accuracy under posterior drift. 
% Notably, DI-ERM also surpasses the previous state-of-the-art reported by \citet{deng2023you}.

\begin{table}[h]
    \centering
    % \caption{
    % Test accuracy on the sentiment disagreement dataset. We benchmarked several popular DG methods. Incorporating annotator profiles ($M$) through DI-ERM yields a dramatic improvement over pooling-ERM, reflecting the importance of modeling annotator-specific posterior drift. 
    % In particular, DI-ERM nearly doubles accuracy compared to pooling-ERM and surpasses the previous state-of-the-art by \citet{deng2023you}.
    % }

    \caption{
    Test accuracy on the sentiment disagreement dataset. Previous DG methods had not been benchmarked on this dataset; we re-implemented them for comparison under our setup. Incorporating annotator profiles ($M$) through DI-ERM yields a dramatic improvement over pooling-ERM, reflecting the importance of modeling annotator-specific posterior drift. 
    % In particular, DI-ERM nearly doubles accuracy compared to pooling-ERM. 
    % and surpasses the previous state-of-the-art by \citet{deng2023you}.
    }

    % \resizebox{0.65\linewidth}{!}{
    % \begin{tabular}{@{}ccc@{}}
    %     \toprule
    %     \textbf{Algorithm} & \textbf{Model} & \textbf{Test Avg Acc} \\ \midrule
    %     pooling-ERM  & BERT & 49.1 $\pm$ 0.4 \\
    %     DI-ERM  & BERT & \textbf{90.5 $\pm$ 0.2}   \\
    %     \bottomrule 
    % \end{tabular}

    \begin{tabular}{@{}lcc@{}}
        \toprule
        \textbf{Algorithm} & \textbf{Model} & \textbf{Test Avg Acc} \\ \midrule
        % pooling-ERM                & BERT           & 49.1 $\pm$ 0.4 \\
        IRM                & BERT           & 48.1 $\pm$ 0.7 \\
        GroupDRO           & BERT           & 49.1 $\pm$ 0.1 \\
        CORAL              & BERT           & 48.4 $\pm$ 0.2 \\
        % AnnEmb (SOTA)      & BERT           & 64.6 $\pm$ 0.8 \\
        \midrule
         Pooling-ERM                & BERT           & 49.1 $\pm$ 0.4 \\
        D3G & BERT & 54.2 $\pm$ 0.4 \\
        DI-ERM & BERT   & \textbf{90.5 $\pm$ 0.2} \\ 
        \bottomrule
    \end{tabular}

    % }
    \label{tab:dierm_annotator}
\end{table}

\subsection{Reviewer-specific Analysis}
\label{subsec: reviewer-spec-sa}

% \paragraph{Reviewer-specific sentiment analysis}
We next examine the WILDS-Amazon Reviews dataset \citep{koh2021wilds}, which captures distributional shifts across reviewers. Here, the input \( x \) is a product review, \( y \in \{1, \dots, 5\} \) is the star rating, \( d \) denotes the reviewer identity, and \( m \) consists of all (unlabeled) reviews written by that reviewer. Once again, because of differences among the rating patterns of reviewers, posterior drift is exhibited. 

The central hypothesis is that a reviewer’s writing style $M = P_{X|D}$ provides a useful signal for predicting their rating behavior $P_{Y|X,D}$. The training set contains 245{,}502 reviews from 1{,}252 reviewers, while the test set consists of 100{,}050 reviews from 1{,}334 unseen reviewers.

To incorporate reviewer context $M$, we randomly sample 20 additional reviews written by the same reviewer and concatenate them with the current review in a prompt format (\Cref{fig:reviewer_prompt} in \Cref{appx: experimental-details}). 
We finetune the \texttt{nomic-embed-text-v1.5} model \citep{nussbaum2025nomic}, which supports 
a longer context window than BERT so that domain information $M$ can fit in.
As summarized in \Cref{tab:dierm_wilds}, DI-ERM outperforms pooling-ERM. Beyond higher average accuracy, DI-ERM also boosts the 10th-percentile accuracy across reviewers—a key robustness metric used on the official leaderboard. 
% With end-to-end fine-tuning, DI-ERM surpassing the best leaderboard result on both metrics.
% (see \Cref{appx: experimental-details} for additional discussion).

\begin{table}[h]
    \centering
    \caption{
    % Sentiment classification performance on Amazon-WILDS with reviewer-specific context.
    % DI-ERM consistently improves over pooling-ERM, both in average accuracy and in 10th-percentile reviewer accuracy—the official leaderboard metric.
    % The top part of the table is quoted from the official 
    % WILDS leaderboard (\url{https://wilds.stanford.edu/}). 
    % DI-ERM exceeds the best result reported on the leaderboard.
    Sentiment classification performance on Amazon-WILDS with reviewer-specific context.
    The first block reports results quoted directly from the official WILDS leaderboard 
    (\url{https://wilds.stanford.edu/}), which uses DistilBERT.
    The second block reports our own experiments using nomic-embed-text-v1.5 model.
    DI-ERM improves over pooling-ERM, both in average accuracy and in 10th-percentile reviewer accuracy—the official leaderboard metric.
    % —and exceeds the best leaderboard result by \citep{yao2022improving}.
    } 
    % \vspace{1em}
    % \resizebox{1\linewidth}{!}{
    % \begin{tabular}{@{}cccc@{}}
    %     \toprule
    %     \textbf{Algorithm} & \textbf{Model} & \textbf{Test Avg Acc} & \textbf{Test 10\% Acc} \\ \midrule
    %     pooling-ERM & nomic-embed-text & 72.2 $\pm$ 1.1 & 55.1 $\pm$ 0.8 \\ 
    %     % Fine-tune: text concat
    %     DI-ERM & nomic-embed-text & \textbf{ 73.4 $\pm$ 0.1 }
    %     & \textbf{ 56.4 $\pm$ 0.8 }  \\ 
    %     \bottomrule 
    % \end{tabular}
    \begin{tabular}{@{}lccc@{}}
        \toprule
        \textbf{Algorithm} & \textbf{Model} & \textbf{Test Avg Acc} & \textbf{Test 10\% Acc} \\ \midrule
        Pooling-ERM                & DistilBERT     & 72.0 $\pm$ 0.1  & 54.2 $\pm$ 0.8 \\
        GroupDRO           & DistilBERT     & 70.0 $\pm$ 0.5  & 53.3 $\pm$ 0.8 \\
        CORAL              & DistilBERT     & 71.1 $\pm$ 0.3  & 52.9 $\pm$ 0.8 \\
        IRM                & DistilBERT     & 70.3 $\pm$ 0.6  & 52.4 $\pm$ 0.8 \\
        LISA (SOTA)        & DistilBERT     & 70.7 $\pm$ 0.3  & 54.7 $\pm$ 0.0 \\
        \midrule
        Pooling-ERM & nomic-embed & 72.2 $\pm$ 1.1 & 55.1 $\pm$ 0.8 \\ 
        % Fine-tune: text concat 
        D3G & nomic-embed & 71.8 $\pm$ 0.1 & 54.7 $\pm$ 0.0 \\
        DI-ERM & nomic-embed & \textbf{73.4 $\pm$ 0.1}
        & \textbf{56.4 $\pm$ 0.8 }  \\ 
        \bottomrule
    \end{tabular}

    % }
    \label{tab:dierm_wilds}
\end{table}

% for original train/test split based on different images, it works; but when split based on annotators, it doesn't
% \subsection{Personalized Image Aesthetic Assessment}

% We now test a vision setup where there is sign of posterior drift. We choose Leuven Art Personalized Image Set (LAPIS) dataset \citep{Maerten2025LAPIS}, which consists of images of artworks, where each image is annotated by around 20 participants. The input $x$ is an art image, $y \in [0, 100]$ is a numeric aesthetic score, $d$ denotes the annotator identity and $m$ denotes the personal attributes of the annotator (e.g., art interest). 

% Throughout the experiment, we use a pretained vision encoder (e.g., CLIP \citep{radford2021clip}) to get the feature representation of image $x$. For DI-ERM, we encode the metadata $m$ using a pretrained language model (e.g., DistilBERT) following the prompt in \Cref{fig:pacs_prompt} 

% \TODO{add in the experiments}

% conceptually, this setup is the same as ``annotator disagreement''. Now the input is contains both image (art work) and text (annotator demographics)...

\subsection{Image Classification Across Styles}
\label{subsec:image_dg}
% \paragraph{Image classification across styles}
% We next evaluate our method on the PACS dataset \citep{li2017pacs}, which contains images drawn from four distinct visual styles: $d \in \{ \text{Photo (P), Art Painting (A), Cartoon (C), Sketch (S)} \}$. Each image $x$ belongs to one of seven categories, $y \in {\text{ \{Dog, Elephant, Giraffe, Guitar, Horse, House, Person\} }}$. Domain information is represented by a short text description $m$, such as “a photo” or “a pencil sketch” (see \Cref{fig:pacs_prompt}).

Lastly, we evaluate on a setup where there is little to no posterior drift. This is an empirical verification of the theory in \Cref{sec:di_erm_cov_shift}.
We choose
% next evaluate our method on the 
PACS dataset \citep{li2017pacs}, which contains images from four distinct visual styles (aka, domain $d$): Photo (P), Art Painting (A), Cartoon (C), and Sketch (S). Each image $x$ belongs to one of the seven categories: dog, elephant, giraffe, guitar, horse, house, or person. Domain information $m$ is encoded by a short text description $m$, such as ``a photo'' or ``a pencil sketch'' (\Cref{fig:pacs_prompt} in \Cref{appx: experimental-details}).

This vision task satisfies UBC, since a single classifier based on image $x$ should accurately classify all images across domains. Thus, in line with \Cref{sec:di_erm_cov_shift}, we expect any gains from domain metadata to be due to using a restricted function class.
To implement DI-ERM, we use pretrained image foundation models (e.g., CLIP \citep{radford2021clip}, DINOv2 \citep{oquab2023dinov2}, DINOv3 \citep{simeoni2025dinov3}) to extract visual features from $x$, and encode the domain description $m$ using a pretrained language model (DistilBERT) following the prompt in \Cref{fig:pacs_prompt}. The resulting image and text embeddings are concatenated into a joint representation for classification.

We follow the standard PACS evaluation protocol: training on three domains and testing on the held-out fourth domain, repeated across all domain splits. All encoders are frozen, and linear classifiers are trained on top of that.
% the fixed representations.

As shown in \Cref{tab:pacs_performance}, 
% DI-ERM improves over pooling-ERM in most settings. 
the gains of DI-ERM over pooling-ERM are most pronounced for ViT-small models, while the benefit diminishes for ViT-large foundation models. This pattern aligns with the discussion in \Cref{sec:di_erm_cov_shift}: under UBC, the benefit of DI-ERM decreases as model size grows.

% \vspace{-1em}

\begin{table}[htbp]
    \centering
    \renewcommand{\arraystretch}{1.3}
    \caption{
    % Summary of image classification across styles (PACS). We highlight that DI-ERM outperforms the pooling-ERM for a majority of testing scenarios, highlighting the advantage of leveraging domain information.
    % We also observe that the benefit of DI-ERM decreases when we choose larger models in the CLIP and DINOv2 family, this is partially due to that the model is already too powerful. The gradual decrease of the benefit is consistent with the previous finding in \citet{cho2023promptstyler}.
    Domain generalization results on PACS using models from the CLIP, DINOv2 and DINOv3 families.
    DI-ERM achieves improved accuracy over pooling-ERM, but the gain diminishes for large models, a phenomenon consistent with our theory in \Cref{sec:di_erm_cov_shift}.
    When using large models, both ERM and DI-ERM approach SOTA performance by \citet{cho2023promptstyler}.
    % Gains diminish as model size increases, suggesting that large pretrained models already capture style-invariant features.
    }
    \label{tab:pacs_performance}
    % \vspace{1em}
    % \scalebox{0.8}{
    % \resizebox{0.9\linewidth}{!}{
    % \begin{tabular}{@{}ccc@{}}
    %     \toprule
    %     \textbf{Algorithm} & \textbf{Model} & \textbf{Test Avg Acc} \\ \midrule
    %     pooling-ERM (linear) & \multirow{2}{*}{CLIP: vitb32} & 95.0 $\pm$ 0.0 \\
    %     DI-ERM (linear) & & \textbf{95.4} $\pm$ 0.0 \\
    %     \graymidrule
    %     pooling-ERM (linear) & \multirow{2}{*}{CLIP: vitl14} & 98.6 $\pm$ 0.0 \\
    %     DI-ERM (linear) & & 98.6 $\pm$ 0.0 \\
    %     \midrule 
    %     pooling-ERM (linear) & \multirow{2}{*}{DINOv2: vits14} & 87.6 $\pm$ 0.0 \\
    %     DI-ERM (linear) & & \textbf{89.0} $\pm$ 0.0 \\
    %     \graymidrule
    %     pooling-ERM (linear) & \multirow{2}{*}{DINOv2: vitl14} & 96.1 $\pm$ 0.0 \\
    %     DI-ERM (linear) & & \textbf{96.3} $\pm$ 0.0 \\
    %     \midrule
    %     pooling-ERM (linear) & \multirow{2}{*}{DINOv3: vits16} & 91.7 $\pm$ 0.0 \\ 
    %     DI-ERM (linear) & & \textbf{93.3} $\pm$ 0.0 \\
    %     \graymidrule
    %     pooling-ERM (linear) & \multirow{2}{*}{DINOv3: vitl16} & 95.4 $\pm$ 0.0 \\ 
    %     DI-ERM (linear) & & \textbf{96.2} $\pm$ 0.0  \\
    %     \bottomrule 
    % \end{tabular}
    % }

    % \resizebox{0.9\linewidth}{!}{
    \begin{tabular}{@{}ccc@{}}
        \toprule
        \textbf{Algorithm} & \textbf{Model} & \textbf{Test Avg Acc} \\ \midrule
        PromptStyler (SOTA) & CLIP: ViT-large & 98.6 $\pm$ 0.0 \\
        \midrule
        Pooling-ERM (linear) & \multirow{2}{*}{CLIP: ViT-small} & 95.0 $\pm$ 0.0 \\
        DI-ERM (linear) & & \textbf{95.4 $\pm$ 0.0} \\
        \graymidrule
        Pooling-ERM (linear) & \multirow{2}{*}{CLIP: ViT-large} & 98.6 $\pm$ 0.0 \\
        DI-ERM (linear) & & 98.6 $\pm$ 0.0 \\
        \midrule 
        Pooling-ERM (linear) & \multirow{2}{*}{DINOv2: ViT-small} & 87.6 $\pm$ 0.0 \\
        DI-ERM (linear) & & \textbf{89.0 $\pm$ 0.0} \\
        \graymidrule
        Pooling-ERM (linear) & \multirow{2}{*}{DINOv2: ViT-large} & 96.1 $\pm$ 0.0 \\
        DI-ERM (linear) & & \textbf{96.3 $\pm$ 0.0} \\
        \midrule
        Pooling-ERM (linear) & \multirow{2}{*}{DINOv3: ViT-small} & 91.7 $\pm$ 0.0 \\ 
        DI-ERM (linear) & & \textbf{93.3 $\pm$ 0.0} \\
        \graymidrule
        Pooling-ERM (linear) & \multirow{2}{*}{DINOv3: ViT-large} & 95.4 $\pm$ 0.0 \\ 
        DI-ERM (linear) & & \textbf{96.2 $\pm$ 0.0}  \\
        \bottomrule 
    \end{tabular}
    % }

\end{table}

\section{Conclusions}
This work presents a rigorous theory of domain generalization, and identifies \emph{posterior drift} as the key regime in which domain information at test time can be strictly beneficial. 
% precisely characterizing when and why leveraging domain information at test time is beneficial. 
Empirically, we demonstrate that domain-informed ERM (DI-ERM) outperforms pooling-ERM across several representative scenarios under posterior drift.

A central practical requirement of DI-ERM is access to domain metadata; when such metadata is unavailable, unlabeled data from the target domain offers a natural alternative.
Future work might explore alternative ways of encoding domain information, and a broader range of DG benchmarks. Additionally, it would be interesting to explore 
% the phenomena discussed 
% in the context of 
other performance measures for DG, such as the worst-case test error in \citet{dwork2025many}.

% \newpage

\section*{Acknowledgements}
We thank Gilles Blanchard for helpful discussions and
insights.

\bibliography{refs}
\bibliographystyle{plainnat}

%%%%%%%%%%%%%%%%%%%%%%%%%%%%%%%%%%%%%%%%%%%%%%%%%%%%%%%%%%%%%%%%%%%%%%%%%%%%%%%
%%%%%%%%%%%%%%%%%%%%%%%%%%%%%%%%%%%%%%%%%%%%%%%%%%%%%%%%%%%%%%%%%%%%%%%%%%%%%%%
% APPENDIX
%%%%%%%%%%%%%%%%%%%%%%%%%%%%%%%%%%%%%%%%%%%%%%%%%%%%%%%%%%%%%%%%%%%%%%%%%%%%%%%
%%%%%%%%%%%%%%%%%%%%%%%%%%%%%%%%%%%%%%%%%%%%%%%%%%%%%%%%%%%%%%%%%%%%%%%%%%%%%%%
\newpage
\appendix
% One-column document (no column switch needed)

% \TODO{check if check list is required}

\section{Additional literature review}
\label{appx:lit_review}

\paragraph{ERM is hard to beat.} 
Empirically, \citet{gulrajani2021search} first emphasized that a well-tuned empirical risk minimization (ERM) baseline outperforms many domain generalization (DG) methods on vision benchmarks. Similar patterns were later observed on the WILDS benchmark \citep{koh2021wilds}, and again in the context of federated domain generalization by \citet{bai2024benchmarking}. 

On the theoretical front, \citet{rosenfeld2021risks} and \citet{gouk2024limitations} studied function classes of the form $f: \mathcal{X} \to \mathcal{Y}$ and concluded that, under common assumptions, ERM cannot be fundamentally outperformed (e.g., in terms of minimax risk). The recent work by \citet{dwork2025many} analyzed the optimality of the ``min-max'' ERM with another performance measure of interest.

% Most DG works only takes input $x$ as input during inference time, there are some methods that leverages unlabeled test data. Our framework work casts that as a special case and give conditions when unlabeled data help

\paragraph{Invariant feature learning.} 
A large body of work in domain generalization aims to learn features that are invariant, or transferable, across domains. 
CORAL \citep{sun2016coral} aligns second-order statistics of feature representations across domains. DANN \citep{ganin2016dann} employs adversarial training to encourage domain-invariant features. 
Invariant Risk Minimization (IRM) \citep{arjovsky2019irm} seeks a representation for which a single classifier is simultaneously optimal across all training domains. 
\citet{bui2021exploiting} disentangle domain-invariant and domain-specific features and discard the latter at inference time. 
For more comprehensive overviews, we refer the reader to the surveys by \citet{wang2022estimating} and \citet{zhou2023domain}.

% There has been a large body of works that aims to learn invariant, or transferable, features across the domains. CORAL \citep{sun2016coral} algins second-order statistics of different domains. DANN \citep{ganin2016dann} uses adversarial training to match feature across domains. IRM \citep{arjovsky2019irm} learns a representation such that a single classifier on top of that is simulatenously optimal for all domains. \citet{bui2021exploiting} proposes to disentangal domain-invariant and domain-specific features and remove the latter during inference time. For a more detailed list, we refer to the survey papers by \citet{wang2022estimating,zhou2023domain}.

\paragraph{The use of unlabeled data.} While most DG methods restrict themselves to using only the input \(x\) at inference time, some methods explore the use of unlabeled test-domain data. 
Several DG methods attempt to exploit unlabeled test data to improve generalization \citep{blanchard2011generalizing, muandet2013domain, zhang2021adaptive}. 
A closely related setting is unsupervised domain adaptation (UDA), where unlabeled test data are used to adapt models to the test domain. Unlike DG, UDA assumes access to target-domain data at training time and typically requires learning a separate model per test domain \citep{sun2016coral, ganin2016dann}. 
% UDA methods have shown empirical success on vision benchmarks \citep{sun2016coral, ganin2016dann}. 
% However, a large-scale evaluation by \citet{sagawa2022wilds2} across ten diverse datasets reported mostly negative results—methods using unlabeled data often fail to improve over strong ERM baselines in practice. 

Although promising in principle, the practical benefits of using unlabeled data remain mixed. A large-scale study by \citet{sagawa2022wilds2} evaluating methods across ten diverse datasets found that incorporating unlabeled data frequently failed to improve upon strong ERM baselines. 
These findings reinforce the need for a more precise understanding of when and how unlabeled data can contribute to domain generalization.

Our framework addresses this gap by casting unlabeled data as a special case of auxiliary domain information, and by providing conditions under which such information is expected to improve generalization performance.

\paragraph{Learning Using Privileged Information (LUPI)}

% \TODO{modify this section}

Our formulation of DG also is related to the learning using privileged information (LUPI) framework 
\citep{vapnik2009new,sharmanska2014learning,vapnik2015learning},
% [Vapnik and Vashist, 2009; Sharmanska et al., 2014; Vapnik and Izmailov, 2015], 
where additional side information is available during training. Recent works have explored LUPI in domain adaptation settings \citep{breitholtz2024unsupervised}.

Two key differences distinguish our setting from LUPI:

\begin{itemize}
    \item LUPI typically provides different privileged information for each sample, whereas in our DG setup, samples from the same domain share the same domain metadata.
    \item Crucially, we assume access to domain metadata at test time, while LUPI typically assumes side information is only available during training. This difference is fundamental to our theoretical contributions: LUPI guarantees smaller sample complexity, while our setting enables strictly smaller Bayes risk.
\end{itemize}

\section{A Simple Example of UBC}
\label{sec:cs_eg}

The example below show that even if there is no decision-theoretic gain of DI-ERM under UBC (\Cref{cor:cov_shift}), it may still have practical advantage when considering learning over a restricted function class.

% Yilun: uncomment this to show the figure
\begin{figure*}[htbp]
% \vspace{-0.4cm}
  \centering
    \includegraphics[width=0.7\linewidth]{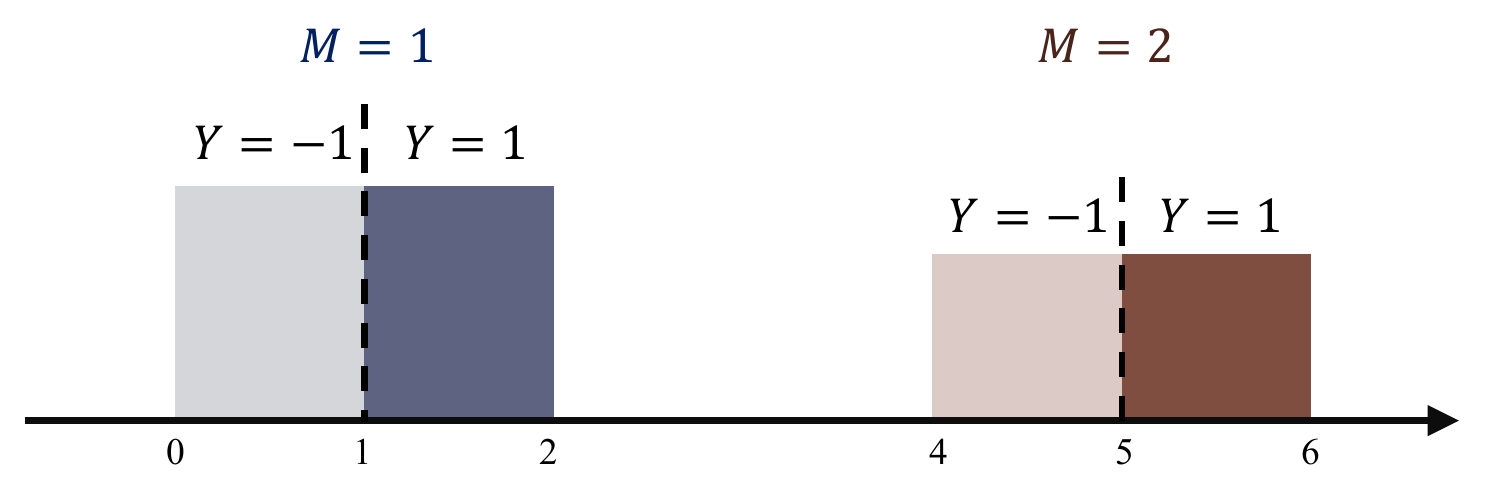}
    \caption{Illustration of Example~\ref{eg:covariate_shift_improve}, where  
    % Domain-informed $\mathcal{F}$ attains zero risk, while pooled $\mathcal{G}$ incurs $\min\{p,1-p\}/2$.
    $\rxg > \rxif$.
    }
    \vspace{-0.4cm}
    \label{fig:example6}
\end{figure*}

\begin{example}
% [Covariate shift without posterior drift]
\label{eg:covariate_shift_improve}

% \textbf{Distribution.} 
Let $P_{XYM}$ be
\begin{align*}
    M\sim\mathrm{Bernoulli}(p), \text{ where } p > 1/2, \\ 
    % \quad 
    \begin{cases}
        M=1: & X\!\sim\!\mathrm{Unif}[0,2],\; Y=\operatorname{sign}(X-1) \\
        M=2: & X\!\sim\!\mathrm{Unif}[4,6],\; Y=\operatorname{sign}(X-5).
    \end{cases}
\end{align*}

Because the supports are disjoint, the pooling and DG Bayes classifier are the same, to be specific
\begin{align*}
    \fx(x) & = \begin{cases}
    \operatorname{sign}(x-1), & x \in[0,2]\\
    \operatorname{sign}(x-5), & x \in[4,6]
    \end{cases}, \\
    % \quad 
    \fxi(x, m) & = \operatorname{sign}(x- 4m + 3)
    \quad \Longrightarrow
    \fx = \fxi
\end{align*}
therefore $\rx = \rxi = 0$.
The model classes are linear classifiers
\begin{align*}
  \mathcal{F} & =\bigl\{f(x,m)=\operatorname{sign}(w^\top x+v^\top m+b)\bigr\}, \\
  % \qquad
  \mathcal{G} & =\bigl\{f(x)=\operatorname{sign}(w^\top x+b)\bigr\}.
\end{align*}
 
$\mathcal{F}$ can realize $\fxi$ with a bias term that depends on $m$, giving
$\rxif=0$.
However, a predictor in $\mathcal{G}$ can only choose a single threshold, and the optimal one is
\begin{align*}
    \fxg(x) = \operatorname{sign}(x-1), \quad \rxg=\frac{\min\{p,1-p\}}{2}.
\end{align*}
Therefore, $\rxg > \rxif$, even though $\rx = \rxi$.
\end{example}

This toy construction mirrors image classification task across different styles: each style
(domain) has a separate support, so the Bayes classifier is the same with or without
$m$, yet $m$ still helps within a restricted model class.
%(e.g., linear probing). 
This is experimentally illustrated in 
% \Cref{subsec:image_dg}
\Cref{sec:experiments}
.

\section{Proofs}

\subsection{Proof of \Cref{prop:risk_hierarchy}}
\label{subsec:risk_hierarchy_proof}

\begin{proposition*}[Risk Hierarchy]
$ \rx \geq \rxi \geq \rxd. $
% \label{prop:risk_hierarchy}
\end{proposition*}

\begin{proof}
Although this result follows from Theorems \ref{thm:dg_improve} and \ref{thm:full_dg_gap}, it can be proved more directly as follows:
    \begin{align*}
        \rx & = \inf_{f: \mc{X} \to \mc{Y}} \expect{X, Y, M, D}{\indicator{f(X) \neq Y}} \\
        & \geq \inf_{f: \mc{X} \times \mc{M} \to \mc{Y}} \expect{X, Y, M, D}{\indicator{f(X, M) \neq Y}} = \rxi \\
        & \geq \inf_{f: \mc{X} \times \mc{M} \times \mc{D} \to \mc{Y}} \expect{X, Y, M, D}{\indicator{f(X, M, D) \neq Y}} \\
%        & = \inf_{f: \mc{X} \times \mc{M} \times \mc{D} \to \mc{Y}} \expect{M, D}{ \expect{X, Y|M, D}{ \indicator{f(X, M, D) \neq Y}} } \\
%        & = \inf_{f: \mc{X} \times \mc{D} \to \mc{Y}} \expect{M, D}{ \expect{X, Y|D}{ \indicator{f(X, D) \neq Y}} } %& \because (X, Y) | M, D = (X, Y) | D 
        & = \inf_{f: \mc{X} \times \mc{D} \to \mc{Y}} \expect{X, Y, M, D}{\indicator{f(X, D) \neq Y}} = \rxd. 
    \end{align*}
To see the final step, note that the $f(x,m,d)$ achieving the minimum in the next-to-last line is $\mathop{\mathrm{argmax}}_k {\mathbb P}(Y=k \mid X=x,M=m,D=d)$. By Assumption \eqref{assump:m}, this is equal to $\mathop{\mathrm{argmax}}_k {\mathbb P}(Y=k \mid X=x,D=d)$, which achieves the minimum in the last line.
\end{proof}

\subsection{Proof of \Cref{thm:dg_improve}}
\label{subsec:dg_improve_proof}

\begin{theorem*}[Risk Reduction from Domain Information] Consider any random triple $(X, Y, M)$, where $Y$ is discrete.
Then
\begin{align*}
    \expect{X, M}{ \gamma(X, M) \indicator{  \fx(X)\neq \fxi(X, M) } } \leq
    % R \Big( \fx(x)\Big) - R \Big( f^*(X, M) \Big) 
    % R^*(X) - R^*(X, M)
    \rx - \rxi
    \leq \expect{X, M}{ \indicator{ \fx(X)\neq \fxi(X, M) } }
\end{align*}
and in particular, 
\begin{align*}
    % R^*(X,M) = R^*(X,D) 
    \fx(X) = \fxi(X, M) \quad \text{ almost surely w.r.t. } P_{XM}
    \implies
    \rx = \rxi.
\end{align*}

% Furthermore, equality on both sides holds if and only if \TODO{seems to be only ``if''}
% \begin{align*}
%     \forall x, m, \ \argmax{k} \prob{Y=k | X = x} = \argmax{k} \prob{Y=k | X=x, M = m}.
% \end{align*}

% \label{thm:dg_improve}
    
\end{theorem*}

\begin{proof}
    The gap in the two risks can be expressed as
    \begin{align*}
        \rx & - \rxi \\
        % = \ \expect{X, M}{ \max_k \prob{Y=k|X, M} } - \expect{X}{ \max_k \prob{Y=k|X=x} } \\
        = & \ \expect{X}{ \expect{M|X}{ \prob{Y= \fxi(X, M) | X, M} } } - \expect{X}{ \prob{Y=\fx(X)| X} } \\
        = & \ \expect{X}{ \expect{M|X}{ \prob{Y= \fxi(X, M) | X, M} } } - \expect{X}{ \expect{M|X}{ \prob{Y=\fx(X)| X, M} } } \\
        = & \ \expect{X}{ \expect{M|X}{ \prob{Y= \fxi(X, M) | X, M} - \prob{Y=\fx(X)| X, M} }  }
    \end{align*}
    % For every $x$ and $m$,

    Next, we analyze the expression in a pointwise manner (aka, for every $x$ and $m$).
    
    Notice that for any $x, m$, if $\fxi (x,m) = \fx (x)$, then the pointwise difference of the conditional probabilities inside the expectation above must be zero.
    
    Whereas if they disagree, then it must hold that
    \begin{align*}
        \prob{Y= \fxi(X, M) | X, M} & = \max_k \prob{Y =  k |X,M},
    \end{align*}
    and
    \begin{align*}
        \prob{Y=\fx(X)| X, M} & \le \operatorname{2nd} \max_k \prob{Y =  k |X,M}.
    \end{align*} 
    
    Thus concluding the two scenarios, the lower bound follows that
    \begin{align*}
        \gamma(x, m) \indicator{\fx(x)\neq \fxi(x, m) } 
        &\leq \prob{Y=\fxi(x, m)|X=x, M=m} - \prob{Y=\fx(x)|X=x, M=m}. 
        % \\
        % &\leq \indicator{\fx(x)\neq f^*_{\mathrm{DI}}(x,m) }.
    \end{align*}
    The upper bound can be bounded by the difference of the two optimal classifiers
    \begin{align*}
        \prob{Y=\fxi(x, m)|X=x, M=m} - \prob{Y=\fx(x)|X=x, M=m}
        &\leq \indicator{\fx(x)\neq f^*_{\mathrm{DI}}(x,m) }.
    \end{align*}
    
    % The inequalities in the theorem statement now follow.

    Taking the expectation over $X$ and $M$, we get the upper and lower bounds in the theorem statement
    \begin{align*}
        \expect{X, M}{ \gamma(X, M) \indicator{  \fx(X)\neq \fxi(X, M) } } \leq
        % R \Big( \fx(x)\Big) - R \Big( f^*(X, M) \Big) 
        % R^*(X) - R^*(X, M)
        \rx - \rxi
        \leq \expect{X, M}{ \indicator{ \fx(X)\neq \fxi(X, M) } }
    \end{align*}

    From the (pointwise) lower and upper bound above, we can  get the sufficient condition that 
    \begin{align*}
        % R^*(X,M) = R^*(X,D) 
        \fx(X) = \fxi(X, M) \quad \text{ almost surely w.r.t. } P_{XM}
        \implies
        \rx = \rxi.
    \end{align*}
    
    % the we have
    % \begin{align*}
    %     \expect{X}{ \expect{M|X}{ \gamma(x, m) \indicator{i^*(x) \neq i^*(x, m) } }  } 
    %     \leq
    %     R \left( \fx(x)\right) - R \left( f^*_{\mathrm{DI}}(x,m) \right)
    %     \leq \expect{X}{ \expect{M|X}{ \indicator{i^*(x) \neq i^*(x, m) } }  }
    % \end{align*}

    % \TODO{Be careful about how to break the tie, the easiest way is to assume $\argmax{k} \prob{Y=k | X=x, M = m}$ is a singleton.}
\end{proof}

\subsection{Proof of \Cref{prop:dg_inf_lower_bound}}
\label{subsec:dg_inf_lower_bound_proof}

\begin{proposition*}
    \begin{align*}
      \inf_{P_{XYMD} \in \Pi(\gamma, \epsilon)} \Big[  \rx - \rxi \Big] \geq  \frac{\gamma \cdot \epsilon }{2}
    \end{align*}
    % \label{prop:dg_inf_lower_bound}
\end{proposition*}

\begin{proof}
    From the lower bound in \Cref{thm:dg_improve}, we have
    \begin{align*}
        \rx - \rxi
        & \geq \expect{X, M}{ \gamma(X, M) \indicator{  \fx(X)\neq \fxi(X, M) } } & \because \text{ \Cref{thm:dg_improve} }     \\
        & \geq \gamma \ \expect{X, M}{\indicator{  \fx(X)\neq \fxi(X, M) } } & \because \text{margin assumption in } \Pi(\gamma, \epsilon) \\
        & = \gamma \ \expect{X}{ \expect{M|X}{ \indicator{  \fx(X)\neq \fxi(X, M) } } } 
    \end{align*}
    Now we will show that
    \begin{align*}
        % \forall X, \quad
        % \expect{M|X}{ \indicator{  \fx(x)\neq f^*(X, M) } } \geq \frac{1}{2} \ \expect{M, M'|X}{ \indicator{  f^*(X, M) \neq f^*(X, M') } },
        \forall x, \quad
        \expect{M|X=x}{ \indicator{  \fx(x)\neq \fxi(x, M) } } \geq \frac{1}{2} \ \expect{M, M'|X=x}{ \indicator{  \fxi(x, M) \neq \fxi(x, M') } },
    \end{align*}
    where 
    \begin{align*}
        M, \ M' \stackrel{\text{i.i.d.}}{\sim} P_{M|X=x}.
    \end{align*}
    Let's examine the two terms. Fix $x$, denote
    \begin{align*}
        \pi_k(x) = \prob{\fxi(x, M) = k | X = x  },
    \end{align*}
    and note that the randomness comes from $M$.
    
    Then for any $x$,
    \begin{align*}
        \expect{M, M'|X=x}{ \indicator{  \fxi(x, M) \neq \fxi(x, M') } } 
        & = \prob{\fxi(x, M) \neq \fxi(x, M') | X = x  } \\
        & = \sum_k \prob{\fxi(x, M) = k, \fxi(x, M') \neq k | X = x} \\ 
        & = \sum_k \pi_k(x) \left(1 - \pi_k(x) \right) \\
        & = 1 - \sum_k \pi_k(x)^2.
    \end{align*}
    Now assume $\fx(x)= k_0$. Then 
    \begin{align*}
        \expect{M|X=x}{ \indicator{  \fx(x)\neq \fxi(x, M) } } 
        & = \expect{M|X=x}{ \indicator{  \fxi(x, M) \neq k_0 } } \\
        & = 1 - \pi_{k_0}(x)
        % \\
        % & = \frac{1}{2} \cdot 2 \cdot \left( 1 - \pi_{k_0} \right)  \\
        % & \geq \frac{1}{2} \cdot (1 + \pi_{k_0})  \cdot \left( 1 - \pi_{k_0} \right)
    \end{align*}
    Notice that
    \begin{align*}
        1 - \sum_k \pi_k(x)^2 
        & \leq 1 - \pi_{k_0}^2 & \textbf{``$=$'' when }  \pi_{k_0} = 1 \\
        & = (1+ \pi_{k_0})(1-\pi_{k_0}) \\
        & \leq 2 (1-\pi_{k_0}) & \textbf{``$=$'' when } \pi_{k_0} = 1.
    \end{align*}
    Then
    \begin{align*}
        \expect{M|X=x}{ \indicator{  \fx(x)\neq \fxi(x, M) } } \geq \frac{1}{2} \ \expect{M, M'|X=x}{ \indicator{  \fxi(x, M) \neq \fxi(x, M') } }.
    \end{align*}
    Integrating over $x$, we have
    \begin{align*}
        \expect{X}{ \expect{M|X}{ \indicator{  \fx(x)\neq \fxi(X, M) } } } 
        & \geq \frac{1}{2} \expect{X}{ \expect{M, M'|X}{ \indicator{  \fxi(X, M) \neq \fxi(X, M') } } } \\
        & = \frac{1}{2} P_{X, M, M'} \Big( \fxi \big(X, M \big) \neq \fxi \big(X, M' \big)   \Big) \\
        & \geq \frac{1}{2} \epsilon. \qquad \because \text{by definition of } \Pi(\gamma, \epsilon)
    \end{align*}
\end{proof}

\subsection{Proof of \Cref{thm:full_dg_gap}}
\label{subsec:full_dg_gap_proof}

\begin{theorem*}
    Let 
    \begin{align*}
        \widetilde{\gamma}(x,d) := 
        \max_k \prob{Y=k|X=x, D=d} - \operatorname{2nd} \max_k \prob{Y=k|X=x, D=d}
        % \max_{y} P_{Y|X,D}(y | x,d) - 2-max_{y} P_{Y|X,D}(y|x,d)
    \end{align*}
    Then 
    \begin{align*}
            \expect{X, D, M}{ \gamma(X, D) \indicator{  \fxd(X,D)\neq \fxi(X, M) } } \leq
    % R^*(X, M) - R^*(X, D)
    \rxi - \rxd
    \leq \expect{X, D, M}{ \indicator{ \fxd(X,D)\neq \fxi(X, M) } }
    \end{align*}
    and in particular, 
    \begin{align*}
        % R^*(X,M) = R^*(X,D) 
        \fxd(X,D) = f^*_{\mathrm{DI}}(X, M) \quad \text{ almost surely w.r.t. } P_{XMD}
        \implies
        \rxd = \rxi.
    \end{align*}
\end{theorem*}

\begin{proof}
    \begin{align*}
        %R^*(X, M) - R^*(X, D)
        \rxi - \rxd
        & = \expect{X, Y, D, M}{ \indicator{Y \neq \fxi(X, D)} } - \expect{X, Y, D, M}{ \indicator{Y \neq \fxd(X, M)} } \\
        & = \expect{X, D, M}{ \prob{Y=\fxd(X, M)} - \prob{Y=\fxi(X, D)} | X, D, M } 
    \end{align*}
    By Assumption \eqref{assump:m}:
    \begin{align*}
        Y|X, D, M = Y|X, D.
    \end{align*} 
    Then for every $x$, $d$ and $m$,
    \begin{align*}
        &\prob{Y=\fxd(x, m) | X=x, D=d, M=m} - \prob{Y=\fxi(x,d)| X=x, D=d, M=m} \\ 
        &=\prob{Y=\fxd(x, m) | X=x, D=d} - \prob{Y=\fxi(x,d)| X=x, D=d} \\
        &\geq  \widetilde{\gamma}(x, d) \indicator{f^*(x, m) \neq f^*(x, d)} .
    \end{align*}
    In the other direction,
    \begin{align*}
         &\prob{Y=\fxd(x, m) | X=x, D=d, M=m} - \prob{Y=\fxi(x,d)| X=x, D=d, M=m} \\ 
        & \leq \indicator{f^*(x, m) \neq f^*(x, d)} .
    \end{align*}
    Integrating over $x, d, m$, we get the lower and upper bounds.

    From the lower and upper bound, we can directly get the sufficient condition
    \begin{align*}
        % R^*(X,M) = R^*(X,D) 
        \fxd(x,d) = f^*_{\mathrm{DI}}(x,m) \quad \text{ almost surely w.r.t. } P_{XMD}.
        \implies
        \rxd = \rxi 
    \end{align*}
\end{proof}

\subsection{Proof of \Cref{prop:risk_hierarchy_class}}

\begin{proposition*}
% [Risk Hierarchy Under Covariate Shift]
% \label{prop:risk_hierarchy_class}
For function classes $\{\mathcal{F}_k\}$ and $\{\mathcal{G}_k\}$ satisfying Assumption~\ref{ass:di_pool_class}. Under UBC, 
% the following 
% %risk hierarchy 
% holds:
\begin{equation}
    R^*_{\mathrm{pool},\mathcal{G}_k} := \inf_{g \in \mathcal{G}_k} R(g)  \geq 
    R^*_{\mathrm{DI},\mathcal{F}_k} := \inf_{f \in \mathcal{F}_k} R(f), \quad \forall k, 
\end{equation}
and
\begin{equation}
    \lim_{k \to \infty} R^*_{\mathrm{pool},\mathcal{G}_k}  = \lim_{k \to \infty} R^*_{\mathrm{DI},\mathcal{F}_k}
    % \label{eqn:cs_asymptotic}
% \end{align*}
\end{equation}

\end{proposition*}

\begin{proof}
    The first line follows from the first assumption of Assumption~\ref{ass:di_pool_class}, where every function $g \in \mathcal{G}_k$ is also realizable in $\mathcal{F}_k$.

    Therefore, we have
    \begin{align*}
    R^*_{\mathrm{pool},\mathcal{G}_k}
        \geq 
        R^*_{\mathrm{DI},\mathcal{F}_k} 
        \geq  
        \rxi 
        \quad \forall k
    \end{align*}

    Take the limit w.r.t. $k$ 

    \begin{align*}
    \lim_{k \to \infty} R^*_{\mathrm{pool},\mathcal{G}_k}
        \geq 
        \lim_{k \to \infty} R^*_{\mathrm{pool},\mathcal{G}_k}
        \geq 
        \rxi 
    \end{align*}

    Recall from the universal approximation property of function class $\mathcal{G}_k$, we have
    \begin{align*}
        \lim_{k \to \infty} R^*_{\mathrm{pool},\mathcal{G}_k} = \rx.
    \end{align*}
    What's more, under UBC assumption, we have
    \begin{align*}
        \rx = \rxi.
    \end{align*}
    Then it must holds that 
    \begin{align*}
        \lim_{k \to \infty} R^*_{\mathrm{pool},\mathcal{G}_k}  = \lim_{k \to \infty} R^*_{\mathrm{DI},\mathcal{F}_k}
    \end{align*}

\end{proof}

\section{Experimental details}
\label{appx: experimental-details}

This section provides additional details on our experimental setup, models, and performance comparisons.
Unless otherwise specified, all models used for fine-tuning are implemented using publicly available checkpoints (e.g., via Huggingface, Pytorch, or official Github repo). 
For linear probing experiments, we extract feature representations using pre-trained transformers and train downstream classifiers with \texttt{scikit-learn}, using either logistic regression or multilayer perceptrons (MLPs). 
% To ensure reproducibility, all results are generated with a fixed random seed $42$.

The following subsections follows the same structure as \Cref{sec:experiments}, while providing additional details and full tables.

\subsection{Annotator disagreement}

\paragraph{Fine-Tuning.}
We fine-tune the \texttt{bert-base-uncased} model and benchmark DI-ERM against other domain generalization methods. For DI-ERM, we concatenate the sentence $x$ with the annotator profile $m$ using the text prompt shown in \Cref{fig:annotator_prompt}.

All experiments are done on NVIDIA A100 or NVIDIA A40 GPU. We finetune the model with logistic loss for 10 epochs with learning rate $5 \cdot 10^{-5}$.
\Cref{tab:annotator_finetune} reports the results over three trials. Our models consistently outperform prior work, with the best configuration achieving over \textbf{$90 \%$} test accuracy—substantially higher than the previous state-of-the-art reported by \citet{deng2023you}.

\begin{figure}[h!]
    \centering
    \begin{tcolorbox}
    [colback=blue!5, colframe=blue!60!black, arc=4pt, boxrule=0.8pt]
    \texttt{Instruction: Read the following sentence and the annotator's demographic profile 
    and determine how positive or negative the annotator judged the sentence on a 
    1--5 scale (1 = Very negative, 5 = Very positive).}
    
    \medskip
    \texttt{Sentence: [sentence goes here]}
    
    \medskip
    \texttt{Annotator profile: Age \{age\}, Race \{race\}, Hispanic/Latino \{hisp\}, grew up in \{grew\}, 
    currently lives in \{curr\}, region \{region\}, income \{income\}, education \{education\}, 
    employment \{employment\}, living situation \{living\}, politics \{politics\}, gender \{gender\}.}
    
    \medskip
    \texttt{Answer:}
    \end{tcolorbox}
    
    \caption{Text prompt that encodes annotator profile.}
    \label{fig:annotator_prompt}
\end{figure}

% We evaluate our approach on a sentiment classification task characterized by annotator disagreement. The table below compares the performance of various algorithms, including our proposed Domain-informed ERM (DI-ERM), with the results reported by \citet{deng2023you}. 

\begin{table}[h!]
    \centering
    \caption{Test accuracy on the sentiment disagreement dataset (fine-tuning BERT). DI-ERM (ours) achieves the best performance.}
    \vspace{0.5em}
    \begin{tabular}{@{}lll@{}}
        \toprule
        \textbf{Algorithm} & \textbf{Model} & \textbf{Test Avg Acc} \\ \midrule
        Pooling-ERM                & BERT           & 49.1 $\pm$ 0.4 \\
        IRM                & BERT           & 48.1 $\pm$ 0.7 \\
        GroupDRO           & BERT           & 49.1 $\pm$ 0.1 \\
        CORAL              & BERT           & 48.4 $\pm$ 0.2 \\
        AnnEmb (SOTA)      & BERT           & 64.6 $\pm$ 0.8 \\
        D3G & BERT & 54.2 $\pm$ 0.4 \\
        DI-ERM (ours, fine-tune) & BERT   & \textbf{90.5 $\pm$ 0.2} \\ 
        \bottomrule
    \end{tabular}
    \label{tab:annotator_finetune}
\end{table}

\paragraph{Linear/MLP-probing.}
We also evaluate in a frozen-feature setting, where the language model is fixed and a lightweight classifier is trained on top. Here, $x$ is encoded with a pretrained sentiment model (e.g., [CLS] embedding of  DistilBERT checkpoint that fine-tuned on SST-2 dataset), while $m$ is encoded with the DistilBERT base model. The embeddings are concatenated and passed to either a linear or shallow MLP classifier. The classifiers are trained in \texttt{scikit-learn}.

\Cref{tab:annotator_linear} presents the results. DI-ERM consistently outperforms pooling-ERM across different feature extractors.

\begin{table}[h!]
    \centering
    \caption{
    Test accuracy on the sentiment disagreement dataset (frozen feature extractor). DI-ERM consistently outperforms pooling-ERM, and in some settings surpasses the prior state-of-the-art of \citet{deng2023you}.
    We highlight the best performance reported by \citet{deng2023you} (69.77) and our highest score (83.41).
    $\dagger$: Checkpoints used in \citet{deng2023you} were not publicly specified.
    }
    \vspace{1em}

    \resizebox{0.8\linewidth}{!}{
    \begin{tabular}{@{}ccc@{}}
        \toprule
        \textbf{Algorithm} & \textbf{Model} & \textbf{Test Avg Acc} \\ \midrule
        % Logistic Reg.: text only  
        \multirow{3}{*}{AnnEmb \citep{deng2023you}} & BERT$^\dagger$ & 64.61 \\
         & RoBERTa$^\dagger$ & 60.30 \\
         & DeBERTa$^\dagger$ & \underline{69.77} \\
        \midrule
        pooling-ERM (linear) & distilbert-base-uncased-finetuned-sst-2-english & 45.85  \\
        % Logistic Reg.: demo extended 
        DI-ERM (linear) & distilbert-base-uncased-finetuned-sst-2-english & \textbf{46.42}  \\ 
        % \midrule 
        \graymidrule
        % \hdashline  
        % MLP: text only 
        pooling-ERM (MLP) & distilbert-base-uncased-finetuned-sst-2-english & 55.07  \\
        % MLP (256, 128): demo extended 
        DI-ERM (MLP) & distilbert-base-uncased-finetuned-sst-2-english & \textbf{78.45}  \\
        \midrule
        pooling-ERM (linear) & bert-base-multilingual-uncased-sentiment & 43.06  \\
        DI-ERM (linear) & bert-base-multilingual-uncased-sentiment & \textbf{43.94}  \\ 
        \graymidrule
        pooling-ERM (MLP) & bert-base-multilingual-uncased-sentiment & 53.90 \\
        DI-ERM (MLP) & bert-base-multilingual-uncased-sentiment & \underline{\textbf{83.41}}  \\
        % \midrule
        % pooling-ERM (linear) & twitter-roberta-base-sentiment-latest & 46.77  \\
        % DI-ERM (linear) & twitter-roberta-base-sentiment-latest & 46.82 \\ 
        % \graymidrule
        % pooling-ERM (MLP) & twitter-roberta-base-sentiment-latest &  \\
        % DI-ERM (MLP) & twitter-roberta-base-sentiment-latest & 58.99 \\
        \bottomrule 
    \end{tabular}
    }
    \label{tab:annotator_linear}
\end{table}

\subsection{Reviewer-specific analysis}

\paragraph{Fine-Tuning.}
We fine-tune the \texttt{bert-base-uncased} model and benchmark DI-ERM against other domain generalization methods. For DI-ERM, we concatenate each review $x$ with reviewer context $m$, represented by 20 randomly selected reviews from the same reviewer, using the text prompt in \Cref{fig:reviewer_prompt}.

We choose \texttt{nomic-embed-text-v1.5} \citep{nussbaum2025nomic}, which supports 
a context window of up to 8012 tokens (we choose 2048 in our experiments),
% a 2048-token window (compared to 512 for DistilBERT), 
in order to handle the long reviwer context $m$.

All experiments are done on NVIDIA A100 or NVIDIA A40 GPU. We finetune the model with logistic loss for 2 epochs with learning rate $10^{-5}$.
\Cref{tab:wilds_finetune} reports the results over three trials. DI-ERM achieves the best performance, outperforming previously reported methods on the WILDS leaderboard (\url{https://wilds.stanford.edu/}). 

% We fine-tune for three epochs with a learning rate of 2e-5, weight decay of 0.01, a training batch size of 16, and an evaluation batch size of 64.

\begin{figure}[h!]
    \centering
    \begin{tcolorbox}[colback=blue!5, colframe=blue!60!black, arc=4pt, boxrule=0.8pt, width=0.9\linewidth]
    \texttt{Instruction: Classify the current review based on this reviewer's sentiment patterns.}
    
    \medskip
    \texttt{Current Review: [current review goes here]}
    
    \medskip
    \texttt{Reviewer's Historical Reviews:}
    
    \texttt{Review 1: [review\_1] \; | \; Review 2: [review\_2] \; | \; \dots}
    \end{tcolorbox}
    \caption{Text prompt that encodes reviewer writing style}
    \label{fig:reviewer_prompt}
\end{figure}

\begin{table}[h!]
    \centering
    \caption{Reviewer-specific sentiment analysis. 
    DI-ERM (ours) achieves the highest accuracy, outperforming prior state-of-the-art by \citet{yao2022improving}.}
    \vspace{0.5em}
    \begin{tabular}{@{}llll@{}}
        \toprule
        \textbf{Algorithm} & \textbf{Model} & \textbf{Test Avg Acc} & \textbf{Test 10\% Acc} \\ \midrule
        Pooling-ERM                & DistilBERT     & 72.0 $\pm$ 0.1  & 54.2 $\pm$ 0.8 \\
        GroupDRO           & DistilBERT     & 70.0 $\pm$ 0.5  & 53.3 $\pm$ 0.8 \\
        CORAL              & DistilBERT     & 71.1 $\pm$ 0.3  & 52.9 $\pm$ 0.8 \\
        IRM                & DistilBERT     & 70.3 $\pm$ 0.6  & 52.4 $\pm$ 0.8 \\
        LISA (SOTA)        & DistilBERT     & 70.7 $\pm$ 0.3  & 54.7 $\pm$ 0.0 \\
        \midrule
        Pooling-ERM & nomic-embed-text & 72.2 $\pm$ 1.1 & 55.1 $\pm$ 0.8 \\ 
        % Fine-tune: text concat
        D3G & nomic-embed & 71.8 $\pm$ 0.1 & 54.7 $\pm$ 0.0 \\
        DI-ERM (ours) & nomic-embed-text & \textbf{73.4 $\pm$ 0.1 }
        & \textbf{56.4 $\pm$ 0.8 }  \\ 
        \bottomrule
    \end{tabular}
    \label{tab:wilds_finetune}
\end{table}

\paragraph{Linear/MLP-probing.}
We also evaluate in a frozen-feature setting, where the language model is fixed and only a lightweight classifier is trained. Each review $x$ is represented by its [CLS] embedding from a pretrained sentiment model (e.g., DistilBERT fine-tuned on SST-2). For reviewer context $m$, we average the [CLS] embeddings of all reviews written by that reviewer. The concatenated review and reviewer embeddings are then passed to a linear or a shallow MLP classifier implemented in \texttt{scikit-learn}.

\paragraph{Domain2Vec.}
Inspired by \citet{zaheer2017deepset, deshmukh2018domain2vec}, we implement a Domain2Vec-style module to encode reviewer-specific domain information. Given a set of reviews \(\{x_1, x_2, \dots, x_n\} \sim P_{X|D=d}\) written by reviewer \(d\), we learn a mapping
\[
f(\{x_1, x_2, \dots, x_n\}) = \rho\left(\frac{1}{n} \sum_{i=1}^n \phi(x_i) \right),
\]
where \(\phi\) and \(\rho\) are MLPs that map individual feature representations (extracted from pretrained model) to a latent space and then transform the aggregated feature, respectively. The resulting vector is concatenated with the review representation \(x\) to predict its sentiment label \(y\).

\Cref{tab:wilds_linear} shows the result.

\begin{table}[h]
    \centering
    \caption{
    % Summary of Amazon review sentiment analysis. Deep Set/Domain2Vec does not lead to improved performance. For fine-tuning, we select a random 10 sentences from the same reviewer and append to each review, separated by [SEP] token. We run 3 repeated experiments. The 10 \% quantile accuracy is better than the best reported on the WILDS leaderboard [url]. We have also tried other ways of incorporating the reviewer information, like xxx, the results is comparable. 
    Sentiment classification on Amazon-WILDS with reviewer-specific signals.
``Domain2Vec'' denotes reviewer encoding based on a learned mean embedding.
DI-ERM variants consistently outperform pooling-ERM baselines.
    }
    \vspace{1em}
    \resizebox{1\linewidth}{!}{
    \begin{tabular}{@{}cccc@{}}
        \toprule
        \textbf{Algorithm} & \textbf{Model} & \textbf{Test Avg Acc} & \textbf{Test 10\% Acc} \\ \midrule
        % LISA & DistillBERT-base-uncased & 70.7 (0.3) & 54.7 (0.0) \\ 
        % ERM (rand search) & DistillBERT-base-uncased & 72.0 (0.1) & 54.2 (0.8) \\ 
        % ERM (grid search) & DistillBERT-base-uncased & 71.9 (0.1) & 53.8 (0.8) \\ 
        % Group DRO & DistillBERT-base-uncased & 70.0 (0.5) & 53.3 (0.0) \\ 
        % Fish & DistillBERT-base-uncased & 71.7 (0.1) & 53.3 (0.0) \\ 
        % CORAL & DistillBERT-base-uncased & 71.1 (0.3) & 52.9 (0.8) \\ 
        % IRM & DistillBERT-base-uncased & 70.3 (0.6) & 52.4 (0.8) \\ 
        % Reweighted (Label) & DistillBERT-base-uncased & 68.3 (0.9) & 51.6 (0.8) \\ 
        % \midrule 
        % Logistic Reg.: text only  
        Pooling-ERM (linear) & distilbert-base-uncased-finetuned-sst-2-english  & 67.42  & 48.00 \\
        % Logistic Reg.: mean extended
        DI-ERM (linear) & distilbert-base-uncased-finetuned-sst-2-english  & \textbf{68.21}  & 48.00 \\
        \graymidrule
        % MLP (32): text only 
        Pooling-ERM (MLP) & distilbert-base-uncased-finetuned-sst-2-english  & 67.59 & 48.00 \\
        % MLP (64, 32): mean extended
        DI-ERM (MLP) & distilbert-base-uncased-finetuned-sst-2-english  & \textbf{68.28}  & \textbf{49.33} \\ 
        \graymidrule
        DI-ERM (Domain2Vec) & distilbert-base-uncased-finetuned-sst-2-english  & 68.21   & 48.00   \\
        \midrule
        % Logistic Reg.: text only 
        Pooling-ERM (linear) & bert-base-multilingual-uncased-sentiment & 72.14  & 53.33 \\
        % Logistic Reg.: mean extended 
        DI-ERM (linear) & bert-base-multilingual-uncased-sentiment & \textbf{73.22}  & \textbf{54.67} \\ 
        \graymidrule
        Pooling-ERM (MLP) & bert-base-multilingual-uncased-sentiment & 73.01 &	53.33 \\
        % MLP (64, 32): mean extended 
        DI-ERM (MLP) & bert-base-multilingual-uncased-sentiment & \textbf{73.18} & \textbf{55.07} \\
        \graymidrule
        DI-ERM (Domain2Vec) & bert-base-multilingual-uncased-sentiment & 73.19  & 54.67 \\
        % \midrule
        % Fine-tune: text only 
        % pooling-ERM (fine-tuning) & bert-base-uncased & 72.46 $\pm$ & 56.00 \\ 
        % % Fine-tune: text concat
        % DI-ERM (fine-tuning) & bert-base-uncased & \textbf{73.71} $\pm$ & \textbf{57.33}  \\
        % DI-ERM (Domain2Vec) & \\
        \bottomrule 
    \end{tabular}
    }
    \label{tab:wilds_linear}
\end{table}

% \subsection{Personalized Image Aesthetic Assessment}

% \begin{figure}[h!]
%     \centering
%     \begin{tcolorbox}[colback=blue!5, colframe=blue!60!black, arc=4pt, boxrule=0.8pt, width=0.9\linewidth]
%     \texttt{Annotator profile: Age \{age\}, Race \{race\}, Hispanic/Latino \{hisp\}, grew up in \{grew\}, 
%     currently lives in \{curr\}, region \{region\}, income \{income\}, education \{education\}, 
%     employment \{employment\}, living situation \{living\}, politics \{politics\}, gender \{gender\}.}
%     \end{tcolorbox}
%     \caption{Example of style-specific text prompts used as domain descriptions.}
%     \label{fig:lapis_prompt}
% \end{figure}

\subsection{Image classification across styles}

% We use a wide range of models in the CLIP and DINOv2 family.
We evaluate our approach on the PACS benchmark, which contains four visual styles: Photo (P), Art Painting (A), Cartoon (C), and Sketch (S). To assess robustness to style variation, we test a diverse set of models from the CLIP, DINOv2 and DINOv3 families.

For all the experiment we use the text prompt in \Cref{fig:pacs_prompt}
as input to \texttt{DistillBERT}.

\begin{figure}[h!]
    \centering
    \begin{tcolorbox}[colback=blue!5, colframe=blue!60!black, arc=4pt, boxrule=0.8pt, width=0.9\linewidth]
    \texttt{Domain "photo", text prompt: "a photo"}
    
    \medskip
    \texttt{Domain "art painting", text prompt: "an oil painting"}
    
    \medskip
    \texttt{Domain "cartoon", text prompt: "a colorful cartoon"}

    \medskip
    \texttt{Domain "sketch", text prompt: "a pencil sketch"}
    \end{tcolorbox}
    \caption{Example of style-specific text prompts used as domain descriptions.}
    \label{fig:pacs_prompt}
\end{figure}

We use different models to extract the visual features. Then train a logistic regression on top of these features using Scikit-learn with lbfgs solver, the results are deterministic. \Cref{tab:pacs_performance_full} summarizes the results. 

Across most domain shifts, our proposed DI-ERM method consistently outperforms standard pooling-ERM, highlighting the advantage of incorporating domain-specific information into the representation.

Notably, we observe that the performance gains from DI-ERM tend to diminish as model capacity increases. For the largest models (e.g., CLIP ViT-L/14 and DINOv2 ViT-L/14), the improvement is marginal or saturates. This trend is also observed by various empirical works, e.g. \citet{cho2023promptstyler}.

\begin{table}[h!]
    \centering
    \renewcommand{\arraystretch}{1.3}
    \caption{
    % Summary of image classification across styles (PACS). We highlight that DI-ERM outperforms the pooling-ERM for a majority of testing scenarios, highlighting the advantage of leveraging domain information.
    % We also observe that the benefit of DI-ERM decreases when we choose larger models in the CLIP and DINOv2 family, this is partially due to that the model is already too powerful. The gradual decrease of the benefit is consistent with the previous finding in \citet{cho2023promptstyler}.
    Domain generalization results on PACS using models from the CLIP, DINOv2 and DINOv3 families.
    % For each source-to-target transfer (e.g., PAC $\to$ S), we report accuracy averaged over three runs. 
    DI-ERM achieves improved accuracy over pooling-ERM in most configurations, particularly for mid-sized models.
    % Gains diminish as model size increases, suggesting that large pretrained models already capture style-invariant features.
    }
    \label{tab:pacs_performance_full}
    \vspace{1em}
    % \scalebox{0.8}{
    \resizebox{\linewidth}{!}{
    
    \begin{tabular}{@{}ccccccc@{}}
        \toprule
        \textbf{Model} & \textbf{Algorithm} & PAC $\to$ S & ACS $\to$ P & CSP $\to$ A & SPA $\to$ C & \textbf{Test Avg Acc} \\ \midrule
        % Logistic Reg.: text only  
        \multirow{2}{*}{CLIP: vitb32} & Pooling-ERM (linear)  & 86.97 &	99.58 &	95.90 &	97.48	& 94.98 \\
        & DI-ERM (linear) & \textbf{88.06} & \textbf{99.64} & \textbf{96.29} & \textbf{97.48} & \textbf{95.37} \\
        % \arrayrulecolor{gray}\cline{1-7}
        \graymidrule
        \multirow{2}{*}{CLIP: vitb16} & Pooling-ERM (linear) & 90.89 & 99.70 & 97.51 & 98.76 & 96.70 \\
        & DI-ERM (linear)  & \textbf{91.09} & 99.70 & \textbf{97.61} & \textbf{98.76} & \textbf{96.79} \\
        \graymidrule
        \multirow{2}{*}{CLIP: vitl14} & Pooling-ERM (linear) & 95.42 & 99.94 & 99.22 & 99.79 & 98.59 \\
        & DI-ERM (linear) & 95.32 & 99.94 & \textbf{99.32} & 99.79 & 98.59 \\
        \midrule 
        \multirow{2}{*}{DINOv2: vits14} & Pooling-ERM (linear)  & 79.82 & 85.81 & 93.55 & 91.34 & 87.63 \\
        & DI-ERM (linear) & \textbf{80.45} & \textbf{90.00} & \textbf{94.09} & \textbf{91.60} & \textbf{89.04} \\
        % \arrayrulecolor{gray}\cline{1-7}
        \graymidrule
        \multirow{2}{*}{DINOv2: vitb14} & Pooling-ERM (linear) & 87.27 & 95.45 & 97.66 & 94.67 & 93.76 \\
        & DI-ERM (linear) & \textbf{87.35} & \textbf{96.53} & \textbf{98.05} & 94.50 & \textbf{94.11} \\
        \graymidrule
        \multirow{2}{*}{DINOv2: vitl14} & 
        Pooling-ERM (linear) & 92.29 & 96.41 & 98.14 & 97.48 & 96.08  \\
        & DI-ERM (linear) & \textbf{92.42} & \textbf{97.37} & 98.10 & 97.48 & \textbf{96.34} \\
        \midrule
        \multirow{2}{*}{DINOv3: vits16} & 
        Pooling-ERM (linear) & 85.14 &	94.25 &	94.34 &	92.92 &	91.67  \\
        & DI-ERM (linear) & \textbf{86.51} &	\textbf{97.49} &	\textbf{96.04} &	\textbf{93.34} &	\textbf{93.3} \\ 
        \graymidrule
        \multirow{2}{*}{DINOv3: vitb16} & 
        Pooling-ERM (linear) & 93.61 &	79.58 &	94.58 &	97.53 &	91.33  \\
        & DI-ERM (linear) & 93.36 &	\textbf{90.78} &	94.82 &	97.14 &	\textbf{94.03} \\ 
        \graymidrule
        \multirow{2}{*}{DINOv3: vitl16} & 
        Pooling-ERM (linear) & 96.18 &	92.28 &	94.04 &	99.10 &	95.40  \\
        & DI-ERM (linear) & 95.80 &	\textbf{95.75} &	94.14 &	99.10 &	\textbf{96.20} \\
        \bottomrule 
    \end{tabular}
    }
    % \label{tab:selected_algorithm_performance}
\end{table}

%%%%%%%%%%%%%%%%%%%%%%%%%%%%%%%%%%%%%%%%%%%%%%%%%%%%%%%%%%%%%%%%%%%%%%%%%%%%%%%
%%%%%%%%%%%%%%%%%%%%%%%%%%%%%%%%%%%%%%%%%%%%%%%%%%%%%%%%%%%%%%%%%%%%%%%%%%%%%%%

\end{document}